\documentclass[english]{article}
\usepackage[utf8]{inputenc}
\usepackage[T1]{fontenc}
\usepackage{babel}
\usepackage{amsmath}
\usepackage{graphicx}
\usepackage{fancyhdr}
\usepackage{multirow}
\usepackage{hyperref}
\usepackage{longtable}
\usepackage{siunitx}
\usepackage{eurosym}
\usepackage{amssymb}
\usepackage{algorithm2e}
\usepackage{arydshln}
\usepackage{comment}

\newcommand{\RN}[1]{%
  \textup{\uppercase\expandafter{\romannumeral#1}}%
}

\begin{document}

\title{Enhanced spatio-temporal electric load forecasts using less data with active deep learning}

\author{Arsam Aryandoust\textsuperscript{1{*}}, Anthony Patt\textsuperscript{1}, Stefan Pfenninger\textsuperscript{1}}

\maketitle
\thispagestyle{fancy}

1. Climate Policy Lab, Department of Environmental Systems Science, Swiss Federal Institute of Technology (ETH Zurich) *corresponding author: Arsam Aryandoust (arsama@ethz.ch), CHN J 72.2, Universitaetsstrasse 22, 8006 Zurich, Switzerland \\


\begin{abstract}
An effective way to oppose global warming and mitigate climate change is to electrify our energy sectors and supply their electric power from renewable wind and solar. Spatio-temporal predictions of electric load become increasingly important for planning this transition, while deep learning prediction models provide increasingly accurate predictions for it. The data used for training deep learning models, however, is usually collected at random using a passive learning approach. This naturally results in a large demand for data and associated costs for sensors like smart meters, posing a large barrier for electric utilities in decarbonizing their grids. Here, we test active learning where we leverage additional computation for collecting a more informative subset of data. We show how electric utilities can apply active learning to better distribute smart meters and collect their data for more accurate predictions of load with about half the data compared to when applying passive learning.
\end{abstract}

\newpage

\noindent
An effective way to mitigate climate change is to electrify our energy sectors, and supply their electricity from renewable wind and solar, which are highly fluctuating and uncertain sources of energy \cite{Patt2015, IPCC2018, IPCC2021}. Planning and operating electricity grids under these uncertainties increasingly requires fine-grained and accurate predictions of electric load across very short to long time windows \cite{Hahn2009, Soliman2010}. Among the different types of electric load forecasts that are performed \cite{Alfares2002, Nti2020}, spatio-temporal predictions have gained increasing importance \cite{Shi2017, Tascikaraoglu2018, Severiano2021}; they predict load for times and places, in which we do not have detailed information about electric load profiles in our grids, and operate these as black boxes \cite{Willis2002}. 

While remotely sensed data from e.g. satellite imagery is increasingly easy to access for making spatio-temporal predictions \cite{Rolf2021, Burke2021}, ground truth electric load data remains difficult and expensive to collect \cite{Melo2014}. One reason for this is that electric utilities are limited in the number of smart meters they can place to collect consumption data by social, financial and technical barriers \cite{Milam2014, Sovacool2021}. Another reason is that utilities are further limited in the amount of data they can query from each meter in real-time by constrains like data communication bandwidths and privacy concerns of consumers, something known as the velocity constraint of data \cite{Kezunovic2013, Yu2015}. Figure 1 shows the current state of global smart meter adoption. For regions with a yet low adoption of smart meters like in most of Africa and Latin America, we mainly want to know where to place new smart meters first so as to make the best possible predictions of electric load for yet unsensed parts of our grids, and effectively use our budget for installing new smart meters. For regions with a medium to high adoption of smart meters like in some parts of Asia and Europe, we further want to know when to query data from which smart meter so as to best utilize our measured data for making accurate predictions of load without exceeding our data velocity constraints. 

Recent advances in artificial intelligence (AI) and machine learning (ML) research have created a growing interest in understanding how AI \cite{Stein2020} and ML \cite{Rolnick2019} can help us tackle such climate change related prediction problems. AI is used in 90\% of recently proposed load forecasting algorithms with deep learning (DL) models making up the largest share of these with 28\% \cite{Nti2020}. The default method of choice for training these DL models is passive learning. In passive learning, the data used for training a prediction model is collected from a large pool of candidate data points at random, which naturally results in a large demand for data and sensors in remote sensing applications. In contrast to this, an increasingly emerging approach, known as active learning, leverages additional computation to assess the information content of data points prior to collecting these, such that we can collect the most useful data only and reduce our overall demand for data and sensors \cite{Settles2010, Dasgupta2011, Kumar2020}. Appendix 1 provides a more detailed description of how AI, ML and DL encompass each other, and how active learning has emerged to solve problems associated with passive learning.

Although active learning is well studied across many theoretical use cases and domains, its application for solving important practical problems like collecting data for spatio-temporal predictions of electric load remains poorly explored. As one of the first applications of active learning in a related problem domain, Kuo et al. reduce the computational complexity of predicting electricity prices by sampling a smaller, more informative subset of training data from a large pool of candidate data points using Gaussian processes \cite{Kuo2015}. Wang et al. increase their model accuracy for time series predictions of electric load compared to existing DL models by using a selector-predictor framework in which the selector samples a subset of similar and correlating load segments from the past to be used as training data by a distinct and continuously updated predictor that consists of an ensemble of DL models \cite{Wang2019}. Zhang et al. reduce their dataset bias and as a consequence also their overall generalization errors for time series predictions of electric load by sampling a training subset from a candidate data pool of past consumption data that is more representative of the entire data population based on reducing an expected error metric \cite{Zhang2021}. This leaves the use of active learning for enhancing prediction accuracy or for saving ground truth electric load data when making spatio-temporal predictions of electric load unexplored. 

In this study, we investigate if active deep learning (ADL) can reduce the demand and associated costs for smart meters and data queries while making reasonably good predictions of electric load in time and space compared to when using passive deep learning (PDL). In particular, we investigate (i) how a dataset of the same size picked with ADL impacts the informativeness of training data and generalization performance compared to when using PDL, (ii) how much data and sensors we can save when using ADL while making as accurate predictions as when using PDL, and (iii) whether the sequence of training data picked with ADL has an impact on generalization performance. We first describe our formulated prediction task and applied ADL method.

\begin{figure}
    \centering
    \includegraphics[height = 18cm]{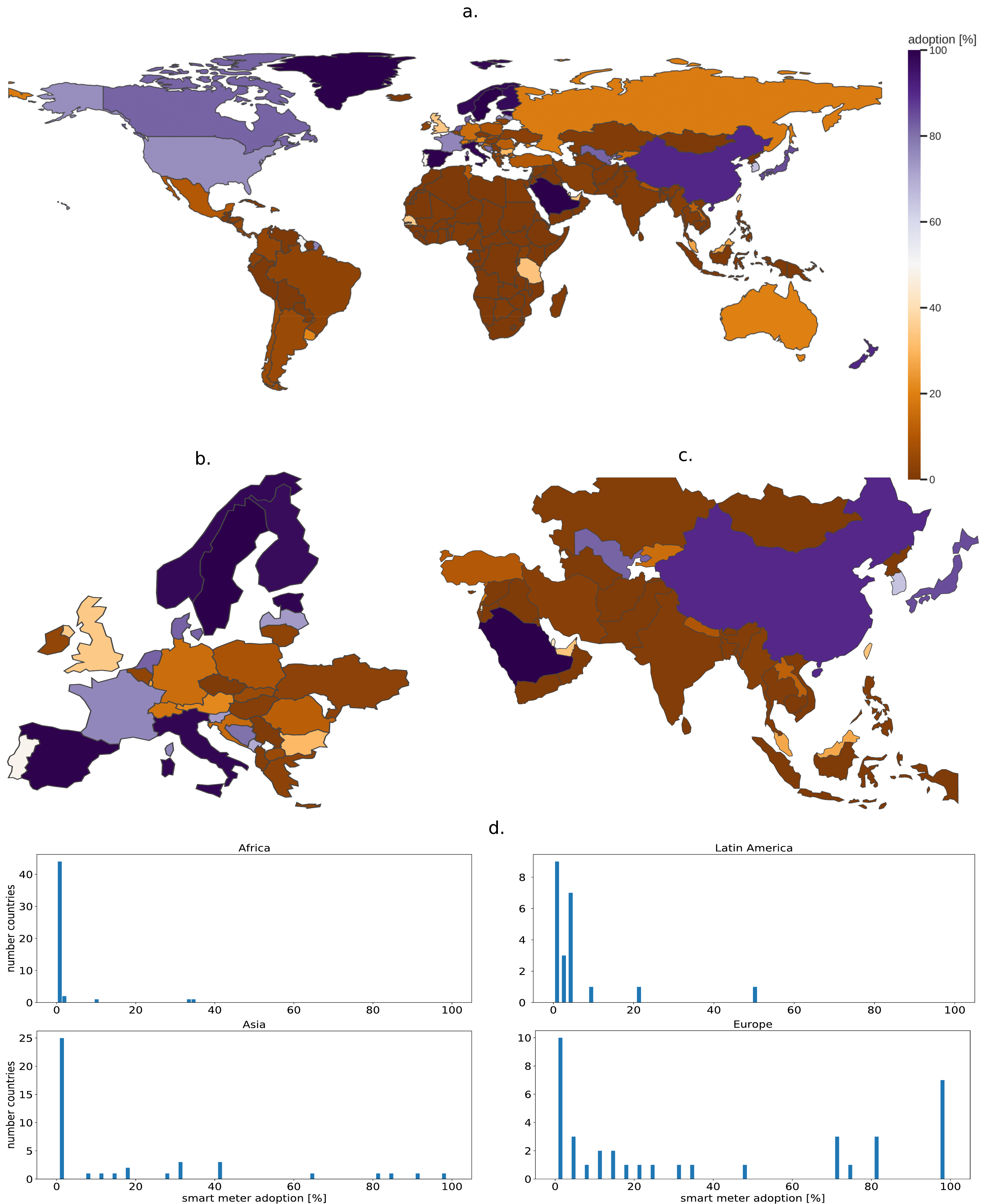}
    \caption{The state of global smart meter adoption (a.). A zoomed in map is further shown for the adoption in European countries (b.), and the adoption in Asian countries (c.). Colors are chosen on a diverging spectrum to better contrast the differences between neighbouring countries. Further, a distribution of national adoption is shown for Africa, Latin America, Asia and Europe (d.).}
\end{figure}

\subsubsection*{Prediction task \& active deep learning method}

Given the aerial image of a building, the meteorological conditions in the region of that building and a time stamp as our features, we want to predict the electric load profile of a building for the next 24 hours in 15 min steps as our labels. Our features (inputs) are all remotely sensed and assumed to be available for every building and point in time at no cost. For every new load profile (output or label) that we collect, we experience some cost and are constrained in the total number of profiles that we can collect by some budget $n_{budget}$. We start with a prediction model that has learnt this relationship for a few buildings and times. Our goal is to collect further ground truth data, i.e. the electric load profiles at different times and buildings, so as to make the best possible predictions for buildings and times, for which we do not have load profiles available, without exceeding $n_{budget}$. Appendix 2 explains the benefits of choosing our data in this constellation in more detail. We run experiments on two different datasets: one containing load profiles from 100, and one from 400 residential, commercial and industrial buildings in Switzerland with diverse sizes, shapes, occupancy and consumption. For each experiment, we randomly select load profiles from 800 (for 100 buildings dataset) and 200 (for 400 buildings dataset) time stamps in 2014 to create a candidate data pool that is of about the same size in each experiment. Figure 2 visualizes the modular DL prediction model architecture that we propose for solving this task; we call it an embedding network.

In each iteration of the ADL algorithm that we apply, we query a batch of candidate data points. First, we encode the features of candidate data points into an embedded vector space using our embedding network prediction model (Figure 2) that is trained on initially available random data points. We then cluster candidate data points based on their vector distances to each other in this encoded space, with the number of clusters being equal to our query batch size. Next, we calculate the distance of the vector of each encoded data point to its cluster center, and query one data point per cluster based on these distances. We test our ADL method for randomized, maximized, minimized and averaged distances of embedded data points to their cluster centers in every queried data batch. We refer to these as our ADL variants. A number of alternative active learning methods exist that we can apply to our data selection task. Appendix 3 describes these in further detail and explains the advantage and potential disadvantage of using the active learning method we apply over existing ones. 

Figure 3 visualizes the difference between data queries with each of our ADL variants. In a first variant of our ADL method, we randomly select data points from each embedded cluster of candidates (Figure 3, a.). In a second variant, we query candidate data points whose embedded feature vectors are furthest away from their cluster centers (Figure 3, b.). We expect to be more uncertain about these points, as they are more likely to be true members of another cluster: we likely explore the data that is close to our decision boundaries, if not outliers, and expect a larger surprise or learning experience from querying labels for these data points. In a third variant, we query labels of data points that are close to their cluster centers, which we expect to be more representative of their clusters and respectively our entire data population (Figure 3, c.). In a fourth variant, we query data points that have the largest distance to the average of distances to cluster centers among all points of the same cluster, which results in a combination of queries alternating between uncertain and representative data points (Figure 3, d.). Each of these ADL variants tries to select a subset of data points from the candidate data pool in a different way that is more informative compared to when selecting these uniformly at random using PDL. The distance of candidate data points to their cluster centers in an embedded vector space is a new metric of information that we propose; we call it embedding uncertainty. 

\begin{figure}
    \centering
    \includegraphics[height = 18cm]{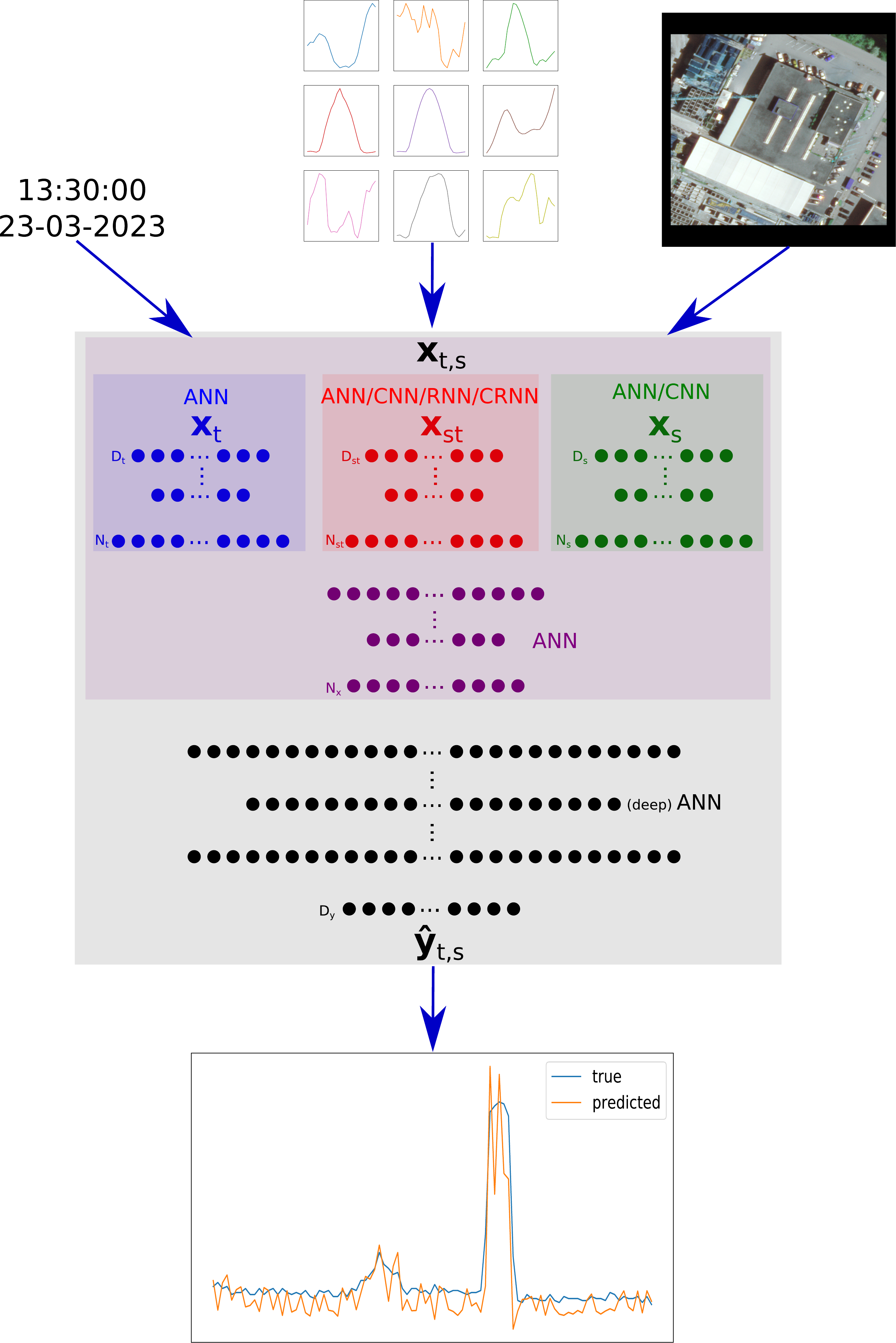}
    \caption{An overview of our embedding network architecture. ANN stand for densely connected, CNN for convolutional, RNN for recurrent, and CRNN for convolutional and recurrent neural network architectures. The inputs or features in this prediction task consist of a time stamp, an aerial image of a building and nine meteorological conditions from the region of that building at the given time. The output or label is the electric consumption of that building at that given time for the next 24 hours in 15 min time steps (96 values).}
\end{figure}

\begin{figure}
    \centering
    \includegraphics[height = 15cm]{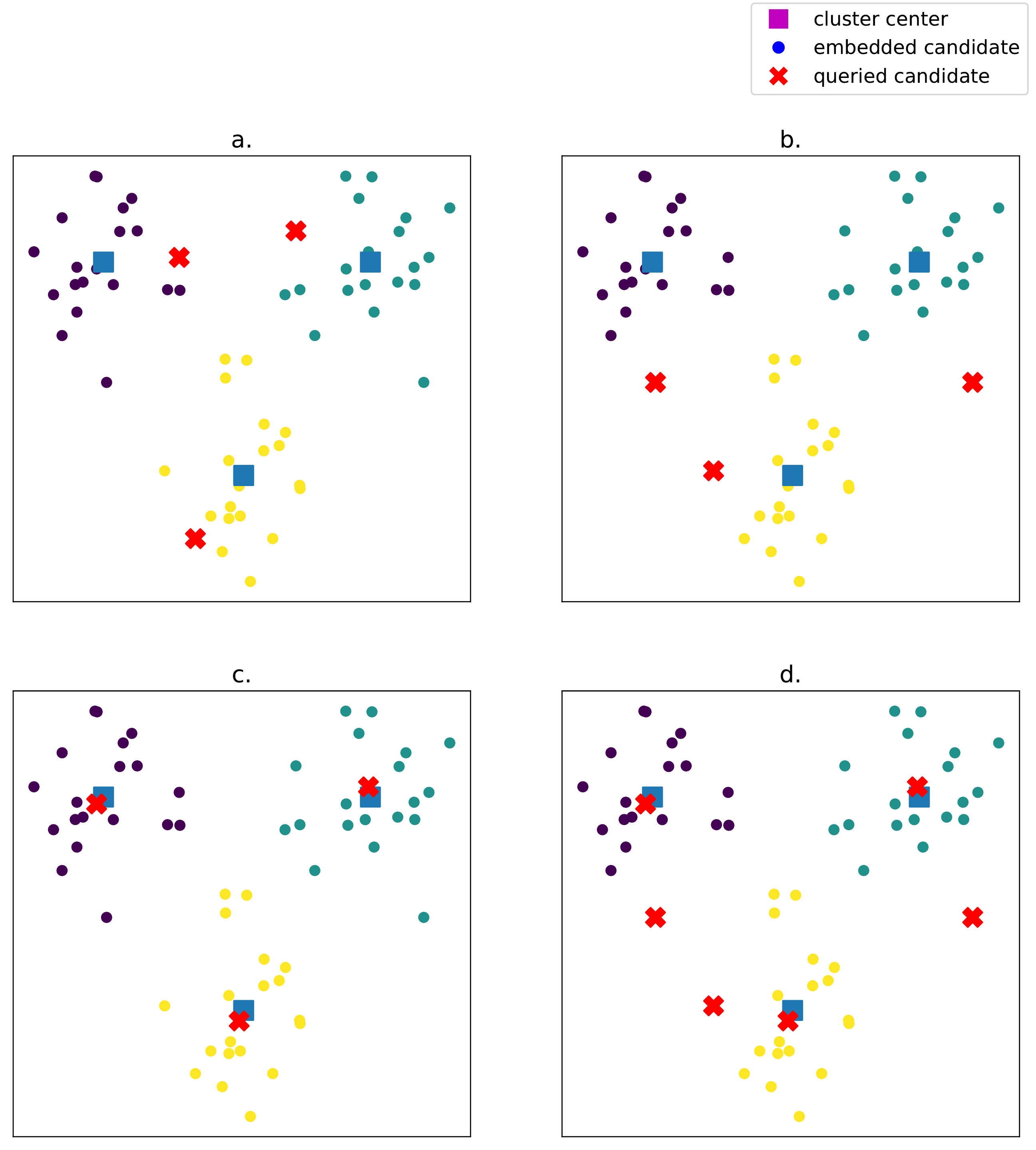}
    \caption{An overview of our ADL variants. Candidate data points are encoded using one of the feature encoders from our embedding network (Figure 2). Here, this is exemplary done for a queried batch size of three and an embedding dimension of two. Points of the same color belong to the same cluster. Squares represent cluster centers. Crosses describe which data points are chosen when a.) randomizing, b.) maximizing, c.) minimizing and d.) averaging the embedding uncertainty of a queried data batch in one iteration of our algorithm; in d.) only one marked point is queried per cluster.}
\end{figure}

We evaluate the performance of our ADL and PDL algorithms for spatial, temporal and spatio-temporal predictions compared to a random forest (RF) benchmark as their common point of comparison. In this context, temporal predictions mean that we predict load profiles for buildings in which a smart meter is placed, but for a time period into the past or future, for which we do not have measured data available; this allows us to compare prediction performance against a distribution shift of our data in time only. Spatial predictions mean that we predict load profiles for buildings in which a smart meter is not placed, but for a time period in which we have load profiles available for other buildings; this allows us to compare prediction performance against a distribution shift of our data in space only. Spatio-temporal predictions respectively refer to the most difficult problem of predicting load profiles for times and buildings, for which we do not have any load profiles available at all; this allows us to compare prediction performance against a distribution shift of our data in both time and space. We refer to these as the different prediction types that we evaluate. 

For each prediction type that we evaluate, we further distinguish between the type of features that we can encode for querying candidate data points. We distinguish between features that are variant in time $\mathbf{x}_t$ (time stamp), space $\mathbf{x}_s$ (building image), and both time and space $\mathbf{x}_{st}$ (meteorological data), as well as the entire feature vector $\mathbf{x}_{t,s}$ which is concatenated from these three vectors.  As the predicted output of our network $\mathbf{\hat{y}}_{t,s}$ represents a vector, i.e. the electric consumption of a building for the next 24 hours in 15-min steps (96 values), we can also use this vector as an embedding of our entire feature vector $\mathbf{x}_{t,s}$. In a further test, we hypothetically use our true labels $\mathbf{y}_{t,s}$ for querying candidate data points so as to see how our proposed metric and ADL variants perform with knowledge about the true distances or similarities of labels that we try to infer from our embedded feature spaces. We refer to these as our ADL variables. 

In all experiments, we remove queried data from the candidate data pool at a rate $\delta \in [0, 1]$, and test this for the two extreme cases of $\delta = 0$ and $\delta = 1$. Removing all queried data from the candidate pool ($\delta = 1$) forces us to exploit our entire data budget and allows us to test how the same number of new data points picked with ADL impacts prediction accuracy compared to when using PDL, and answer our first research question (i). Keeping all queried data in the candidate pool ($\delta = 0$) allows our ADL and PDL algorithms to query the same data multiple times and achieve data and sensor savings; this lets us test how much data and sensors we can save with ADL while making as accurate prediction as compared to PDL, and answer our second research question (ii).

\subsubsection*{Results overview}

When removing queried data from our candidate data pool such that our entire data budget is used ($\delta = 1$), our prediction accuracy increases by up to 37-71\% when using ADL compared to when using PDL. When keeping queried data points in the candidate data pool instead ($\delta = 0$), our demand for data is reduced by up to 41-62\% while achieving an up to 8-45\% higher prediction accuracy, and our demand for sensors can be reduced by up to 12\% while achieving an up to 22\% higher prediction accuracy using ADL compared to when using PDL.

Table 1 contains the numerical results of the experiments that we conduct for each prediction type, ADL variable and ADL variant on a dataset with 400 buildings. We can observe that when removing all queried data points from the candidate data pool ($\delta = 1$), our testing losses are generally lower than when keeping all queried data points in the candidate data pool ($\delta = 0$). We omit experiments for the ADL variables $\mathbf{x}_{t}$ and $\mathbf{x}_{s}$ as these become redundant due to a smaller number of different data points in the candidate data pool than the batch size that we query in each iteration, which either results in choosing the same data points in each iteration or in a random selection of data that is equal to performing PDL. Appendix 4 explains this in further detail, and contains the numerical results for the experiments on the dataset composed of 100 buildings. Figure 4 visualizes exemplar training and validation losses of our experiments.

\begin{table}
    \caption{Numerical results for each prediction type, ADL variable and ADL variant with experiments on a dataset with 400 buildings. The columns named 'data' present what percent of data budget is used; 'sensors' state what percent of sensors are used from the new sensors initially available in the candidate data pool; 'test loss' states the average L2 loss on yet unqueried data; 'accuracy' is calculated as $1 - min(1, \frac{\text{PDL loss}}{\text{RF loss}})$ and $1 - min(1, \frac{\text{ADL loss}}{\text{RF loss}})$.}
    \begin{center}
    \scalebox{0.7}{
        \begin{tabular}{ ccc|c|c|c|c|c|c|c|c| }
         \cline{4-11}
         & & & \multicolumn{4}{ c| }{Removing queried data ($\delta=1$)} & \multicolumn{4}{ c| }{Keeping queried data ($\delta=0$)}\\
         \hline
         \multicolumn{1}{ |c }{prediction type} & ADL variable & ADL variant & data & sensors & accuracy & test loss & data & sensors & accuracy & test loss \\ \hline
          \multicolumn{1}{ |c }{Spatial}& $\mathbf{x}_{st}$ & rnd $d_{c, st}$ & 100\% & 100\% & 57\% & 0.166733 & 72\% & 100\% & 49\% & 0.215845 \\ 
          \multicolumn{1}{ |c }{ }& & max $d_{c, st}$ & 100\% & 88\% & 61\% & 0.150790 & 18\% & 41\% & 26\% & 0.316328 \\ 
          \multicolumn{1}{ |c }{ }& & min $d_{c, st}$ & 100\% & 100\% & 29\% & 0.271004 & 26\% & 99\% & 29\% & 0.303218 \\ 
          \multicolumn{1}{ |c }{ }& & avg $d_{c, st}$ & 100\% & 91\% & 61\% & 0.149194 & 18\% & 40\% & 18\% & 0.349805 \\ 
          \multicolumn{1}{ |c }{ }& $\mathbf{x}_{t,s}$ & rnd $d_{c, (t,s)}$ & 100\% & 100\% & 80\% & 0.077516 & 76\% & 100\% & 79\% & 0.090302 \\ 
          \multicolumn{1}{ |c }{ }& & max $d_{c, (t,s)}$ & 100\% & 100\% & 86\% & 0.054617 & 60\% & 100\% & 67\% & 0.141586 \\ 
          \multicolumn{1}{ |c }{ }& & min $d_{c, (t,s)}$ & 100\% & 100\% & 85\% & 0.058652 & 73\% & 100\% & 73\% & 0.113840 \\ 
          \multicolumn{1}{ |c }{ }& & avg $d_{c, (t,s)}$ & 100\% & 100\% & 84\% & 0.062160 & 67\% & 100\% & 68\% & 0.135954 \\ 
          \multicolumn{1}{ |c }{ }& $\mathbf{\hat{y}_{t,s}}$ & rnd $d_{c, (t,s)}$ & 100\% & 100\% & 92\% & 0.031373 & 48\% & 98\% & 85\% & 0.064173 \\ 
          \multicolumn{1}{ |c }{ }& & max $d_{c, (t,s)}$ & 100\% & 99\% & 91\% & 0.033212 & 38\% & 85\% & 80\% & 0.085099 \\ 
          \multicolumn{1}{ |c }{ }& & min $d_{c, (t,s)}$ & 100\% & 99\% & 91\% & 0.035992 & 49\% & 97\% & 83\% & 0.072779 \\ 
          \multicolumn{1}{ |c }{ }& & avg $d_{c, (t,s)}$ & 100\% & 99\% & 92\% & 0.030670 & 39\% & 87\% & 78\% & 0.093965 \\ 
          \multicolumn{1}{ |c }{ }& $\mathbf{y}_{t,s}$ & rnd $d_{c, (t,s)}$ & 100\% & 99\% & 92\% & 0.029992 & 43\% & 98\% & 44\% & 0.236887 \\ 
          \multicolumn{1}{ |c }{ }& & max $d_{c, (t,s)}$ & 100\% & 95\% & 92\% & 0.029122 & 23\% & 66\% & 33\% & 0.285128 \\ 
          \multicolumn{1}{ |c }{ }& & min $d_{c, (t,s)}$ & 100\% & 96\% & 90\% & 0.037045 & 28\% & 74\% & 23\% & 0.328181 \\ 
          \multicolumn{1}{ |c }{ }& & avg $d_{c, (t,s)}$ & 100\% & 98\% & 92\% & 0.030446 & 25\% & 73\% & 14\% & 0.367300 \\  \hline
          \multicolumn{3}{ |c| }{passive deep learning (PDL) benchmark} & 100\% & 100\% & 39\% & 0.235506 & 80\% & 100\% & 35\% & 0.275150 \\  \hline
          \multicolumn{3}{ |c| }{random forest (RF) baseline} & 0\% & 0\% & 0\% & 0.383450 & 0\% & 0\% & 0\% & 0.425898 \\  \hline
          \multicolumn{1}{ |c }{Temporal}& $\mathbf{x}_{st}$ & rnd $d_{c, st}$ & 100\% & 0\% & 40\% & 0.460572 & 77\% & 0\% & 26\% & 0.207233 \\ 
          \multicolumn{1}{ |c }{ }& & max $d_{c, st}$ & 100\% & 0\% & 55\% & 0.345621 & 21\% & 0\% & 0\% & 0.324328 \\ 
          \multicolumn{1}{ |c }{ }& & min $d_{c, st}$ & 100\% & 0\% & 38\% & 0.479890 & 30\% & 0\% & 9\% & 0.255990 \\ 
          \multicolumn{1}{ |c }{ }& & avg $d_{c, st}$ & 100\% & 0\% & 52\% & 0.372317 & 21\% & 0\% & 0\% & 0.311569 \\ 
          \multicolumn{1}{ |c }{ }& $\mathbf{x}_{t,s}$ & rnd $d_{c, (t,s)}$ & 100\% & 0\% & 81\% & 0.144378 & 76\% & 0\% & 53\% & 0.133416 \\ 
          \multicolumn{1}{ |c }{ }& & max $d_{c, (t,s)}$ & 100\% & 0\% & 84\% & 0.123898 & 55\% & 0\% & 35\% & 0.182810 \\ 
          \multicolumn{1}{ |c }{ }& & min $d_{c, (t,s)}$ & 100\% & 0\% & 80\% & 0.150325 & 71\% & 0\% & 47\% & 0.149987 \\ 
          \multicolumn{1}{ |c }{ }& & avg $d_{c, (t,s)}$ & 100\% & 0\% & 80\% & 0.154796 & 64\% & 0\% & 46\% & 0.150910 \\ 
          \multicolumn{1}{ |c }{ }& $\mathbf{\hat{y}}_{t,s}$ & rnd $d_{c, (t,s)}$ & 100\% & 0\% & 93\% & 0.050963 & 44\% & 0\% & 65\% & 0.099615 \\ 
          \multicolumn{1}{ |c }{ }& & max $d_{c, (t,s)}$ & 100\% & 0\% & 93\% & 0.050067 & 35\% & 0\% & 54\% & 0.128505 \\ 
          \multicolumn{1}{ |c }{ }& & min $d_{c, (t,s)}$ & 100\% & 0\% & 93\% & 0.053345 & 44\% & 0\% & 63\% & 0.104819 \\ 
          \multicolumn{1}{ |c }{ }& & avg $d_{c, (t,s)}$ & 100\% & 0\% & 93\% & 0.050302 & 37\% & 0\% & 62\% & 0.107712 \\ 
          \multicolumn{1}{ |c }{ }& $\mathbf{y}_{t,s}$ & rnd $d_{c, (t,s)}$ & 100\% & 0\% & 92\% & 0.59176 & 39\% & 0\% & 44\% & 0.158606 \\ 
          \multicolumn{1}{ |c }{ }& & max $d_{c, (t,s)}$ & 100\% & 0\% & 91\% & 0.070442 & 22\% & 0\% & 0\% & 0.378247 \\ 
          \multicolumn{1}{ |c }{ }& & min $d_{c, (t,s)}$ & 100\% & 0\% & 89\% & 0.082155 & 26\% & 0\% & 16\% & 0.236202 \\ 
          \multicolumn{1}{ |c }{ }& & avg $d_{c, (t,s)}$ & 100\% & 0\% & 91\% & 0.070920 & 26\% & 0\% & 2\% & 0.276220 \\  \hline
          \multicolumn{3}{ |c| }{passive deep learning (PDL) benchmark} & 100\% & 0\% & 22\% & 0.601375 & 80\% & 0\% & 29\% & 0.200968 \\  \hline
          \multicolumn{3}{ |c| }{random forest (RF) baseline} & 0\% & 0\% & 0\% & 0.770247 & 0\% & 0\% & 0\% & 0.281539 \\  \hline
          \multicolumn{1}{ |c }{Spatio-temporal}& $\mathbf{x}_{st}$ & rnd $d_{c, st}$ & 100\% & 100\% & 48\% & 0.224996 & 72\% & 100\% & 67\% & 0.223128 \\ 
          \multicolumn{1}{ |c }{ }& & max $d_{c, st}$ & 100\% & 93\% & 20\% & 0.348070 & 24\% & 65\% & 45\% & 0.368072 \\ 
          \multicolumn{1}{ |c }{ }& & min $d_{c, st}$ & 100\% & 100\% & 54\% & 0.197638 & 30\% & 100\% & 62\% & 0.253112 \\ 
          \multicolumn{1}{ |c }{ }& & avg $d_{c, st}$ & 100\% & 93\% & 19\% & 0.351947 & 24\% & 66\% & 44\% & 0.374147 \\ 
          \multicolumn{1}{ |c }{ }& $\mathbf{x}_{t,s}$ & rnd $d_{c, (t,s)}$ & 100\% & 100\% & 76\% & 0.103621 & 76\% & 100\% & 77\% & 0.157090 \\ 
          \multicolumn{1}{ |c }{ }& & max $d_{c, (t,s)}$ & 100\% & 100\% & 80\% & 0.086358 & 60\% & 100\% & 66\% & 0.227267 \\ 
          \multicolumn{1}{ |c }{ }& & min $d_{c, (t,s)}$ & 100\% & 100\% & 78\% & 0.093203 & 74\% & 100\% & 73\% & 0.184296 \\ 
          \multicolumn{1}{ |c }{ }& & avg $d_{c, (t,s)}$ & 100\% & 100\% & 81\% & 0.081237 & 66\% & 100\% & 74\% & 0.178249 \\ 
          \multicolumn{1}{ |c }{ }& $\mathbf{\hat{y}}_{t,s}$ & rnd $d_{c, (t,s)}$ & 100\% & 100\% & 91\% & 0.038287 & 46\% & 100\% & 83\% & 0.116017 \\ 
          \multicolumn{1}{ |c }{ }& & max $d_{c, (t,s)}$ & 100\% & 100\% & 92\% & 0.034973 & 39\% & 99\% & 80\% & 0.133851 \\ 
          \multicolumn{1}{ |c }{ }& & min $d_{c, (t,s)}$ & 100\% & 100\% & 91\% & 0.040976 & 48\% & 100\% & 83\% & 0.117153 \\ 
          \multicolumn{1}{ |c }{ }& & avg $d_{c, (t,s)}$ & 100\% & 100\% & 92\% & 0.035021 & 41\% & 100\% & 81\% & 0.127118 \\ 
          \multicolumn{1}{ |c }{ }& $\mathbf{y}_{t,s}$ & rnd $d_{c, (t,s)}$ & 100\% & 100\% & 90\% & 0.043797 & 40\% & 99\% & 75\% & 0.169657 \\ 
          \multicolumn{1}{ |c }{ }& & max $d_{c, (t,s)}$ & 100\% & 94\% & 86\% & 0.060585 & 22\% & 58\% & 13\% & 0.583133 \\ 
          \multicolumn{1}{ |c }{ }& & min $d_{c, (t,s)}$ & 100\% & 98\% & 87\% & 0.058276 & 28\% & 78\% & 52\% & 0.321868 \\ 
          \multicolumn{1}{ |c }{ }& & avg $d_{c, (t,s)}$ & 100\% & 98\% & 88\% & 0.052049 & 26\% & 72\% & 35\% & 0.439464 \\  \hline
          \multicolumn{3}{ |c| }{passive deep learning (PDL) benchmark} & 100\% & 100\% & 55\% & 0.195662 & 80\% & 100\% & 70\% & 0.203910 \\  \hline
          \multicolumn{3}{ |c| }{random forest (RF) baseline} & 0\% & 0\% & 0\% & 0.432625 & 0\% & 0\% & 0\% & 0.673816 \\ 
         \hline
        \end{tabular}
    }
    \end{center}
\end{table}

\begin{figure}
    \centering
    \includegraphics[height = 20cm]{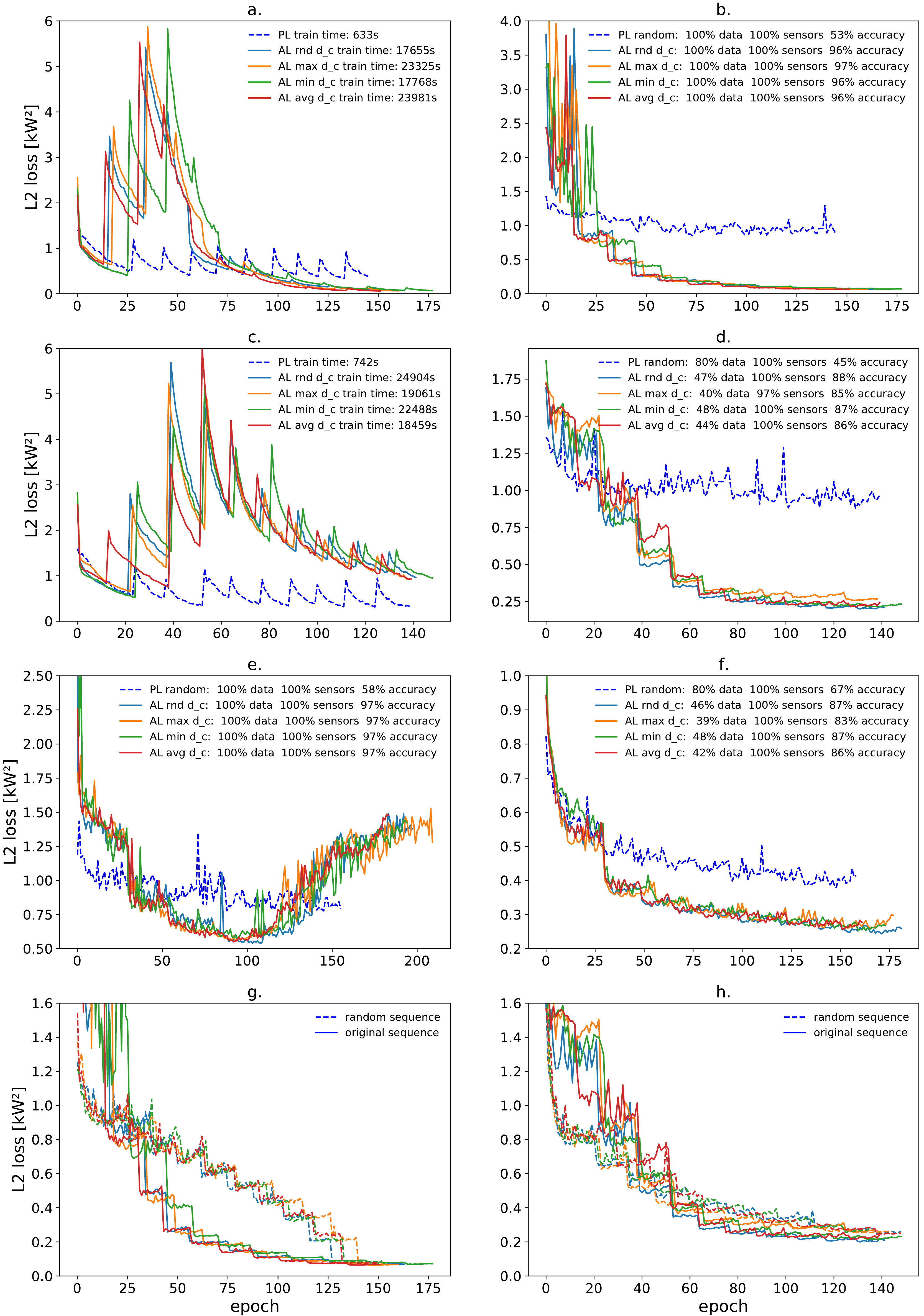}
    \caption{Empirical results for spatio-temporal predictions and ADL variable $\mathbf{\hat{y}}_{t,s}$. a.) and b.): training and validation losses for $\delta=1$. c.) and d.): training and validation losses for $\delta=0$. e.) and f.): validation losses against initial candidates for $\delta=1$ and $\delta=0$. g.) and h.): validation losses with original and randomized query sequence for $\delta=1$ and $\delta=0$.}
    \label{fig:sensor_deployment}
\end{figure}

\subsubsection*{Training \& validation losses for removing all queried data ($\delta=1$)}

In a first set of experiments, we validate predictions on unqueried data points during training and remove all queried data points from the candidate data pool (Figure 4, a. and b.). This allows us to answer our first research question, where we want to investigate how a dataset of the same size picked with ADL impacts the informativeness of training data and generalization performance compared to when using PDL. Training (a.) and validation (b.) losses are visualized for all ADL variants of an exemplar ADL variable compared to PDL (dashed blue line). With each newly queried batch, we can observe that training losses leap up (a.) and validation losses drop (b.). We can further observe that a correlation exists between the magnitude of leaps in training loss and drops in validation loss which are generally larger for ADL than for PDL. Large leaps in training loss indicate that we query data points that are diverse and 'hard to fit to'. They indicate that we implicitly regularize our prediction model by not overfitting to 'easy to learn' patterns in the queried data batch, which lets us generalize better on unqueried data.

\subsubsection*{Training \& validation losses for keeping all queried data ($\delta=0$)}

In a second set of experiments, we validate predictions on unqueried data points during training and keep all queried data points in the candidate data pool (Figure 4 c. and d.). This allows us to achieve data and sensor savings by sampling data points more than once and answer our second research question, where we want to investigate how much data and sensors we can save when using ADL while making as accurate predictions as when using PDL. Training (c.) and validation (d.) losses are visualized for all ADL variants of an exemplar ADL variable compared to PDL (dashed blue line). Unlike in the case of removing all queried candidates, training losses converge to a value that is higher than for PDL (c.). The gap between ADL and PDL validation losses (d.) is smaller compared to when removing queried data from the candidate data pool. We achieve data savings at simultaneously higher prediction accuracy with ADL, which indicates that our ADL method can better learn distribution shifts of our data across time and space compared to PDL.

\subsubsection*{Validation losses against queried and unqueried candidates}

In a third set of experiments, we consider the case of validating accuracy against all initial candidates, which are all queried and unqueried data points during training (Figure 4 e. and f.). These results are not stated in table 1 and do not answer our research questions, but allow us to understand how a bias of our model changes towards yet unqueried candidates for different values of $\delta$, and what effect this has on prediction accuracy. Only validation losses are visualized for removing ($\delta=1$) and keeping ($\delta=0$) all queried data in the candidate pool for this case. For $\delta=1$ (e.), validation losses first decrease and then increase above PDL values towards the last iterations of the algorithm. We create a tendency to forget previously learnt information upon learning new information which results in better predictions on remaining candidates and less accurate predictions on already queried candidates; this is seen from our generally lower testing losses compared to $\delta=0$ and the increase of validation losses towards the last iterations. For $\delta=0$ (f.), our prediction model does not contain a bias and remains resilient against such tendencies; we observe higher testing losses on yet unqueried candidates. We can hence observe that values of $\delta \rightarrow 1$ implicitly cause 'catastrophic inference' or 'forgetting' which increase prediction accuracy on unqueried data and values of $\delta \rightarrow 0$ allow us 'incremental learning'.

\subsubsection*{Importance of ADL query sequence}

In a forth set of experiments, we compare our validation losses during ADL with the case in which we randomize the sequence of queries with the same data during training (Figure 4 g. and h.). This allows us to answer our third research question and see whether purely the information content of queried data matters or whether the query sequence is also meaningful when we perform ADL. For $\delta=1$ (g.), the original ADL sequences converge faster to their final generalization errors than when querying the same data in a random sequence. For $\delta=0$ (h.), our prediction model remains invariant towards the query sequence of candidates. We can hence observe that values of $\delta \rightarrow 1$ can achieve a lower 'regret' in an online learning context than values of $\delta \rightarrow 0$.

\subsubsection*{Discussion}

We investigate (i) how a dataset of the same size picked with ADL impacts the informativeness of training data and generalization performance compared to when using PDL, (ii) how much data and sensors we can save when using ADL while making as accurate predictions as when using PDL, and (iii) whether the sequence of training data picked with ADL has an impact on generalization performance, for predicting electric load profiles of single buildings in time and space. Surprisingly, our empirical results show that we can achieve an even higher prediction accuracy with about half the volume of data when leveraging additional computation for selecting a more informative subset of data with ADL compared to when performing PDL. When using our entire data budget, we achieve an up to 37-71\% higher prediction accuracy (i). Our demand for data can be reduced by up to 41-62\% while achieving an up to 8-45\% higher prediction accuracy, and our demand for smart meters can be reduced by up to 12\% with an up to 22\% higher prediction accuracy (ii). We find that the sequence in which we select data with our ADL method is further useful: training our DL models with data selected in a sequence picked by our ADL method creates a lower regret compared to randomizing the sequence of the same data (iii).

Our findings are consistent with those of previous studies where ADL has led to better predictions of electric load \cite{Wang2019, Zhang2021}, or a higher classification accuracy was achieved with a smaller amount of data points using ADL based on a variant of information entropy \cite{Siddiqui2020}. Nevertheless, our reduced data demand and simultaneously increased prediction accuracy exceeds those observed in previous studies. One reason for this can be larger imbalances in our data, as well as space and time variances of the data distribution that our prediction task involves. These are, however, frequently encountered characteristics of real-world data, which we show ADL is able to tackle to more effectively than PDL.

Our findings can have important implications for the clean energy transition and for mitigating climate change. The ADL method we propose can be used by distribution and transmission system operators for electricity around the world to make more accurate spatio-temporal predictions of electric load using budgets for installing smart meters and streaming their data more effectively: figure 5 shows how our ADL model can point utilities to where to place smart meters next, and when to collect their data for this. For the general reliability and applicability of our findings, our proposed ADL method must be tested on a larger variety of datasets and prediction tasks and be compared against alternative ADL methods (Appendix 3). Further research must also explore how contrastive learning \cite{Chen2020} and domain adaptation \cite{Yue2021, Yang2020, Saito2020} can further reduce data and sensor demand while increasing prediction accuracy prior to applying ADL, and how ADL can be used for spatio-temporal predictions with graph neural networks \cite{Jain2016} to further reduce the computational complexity for training DL models and assessing the informativeness of candidate data points.

\begin{figure}
    \centering
    \includegraphics[height = 20cm]{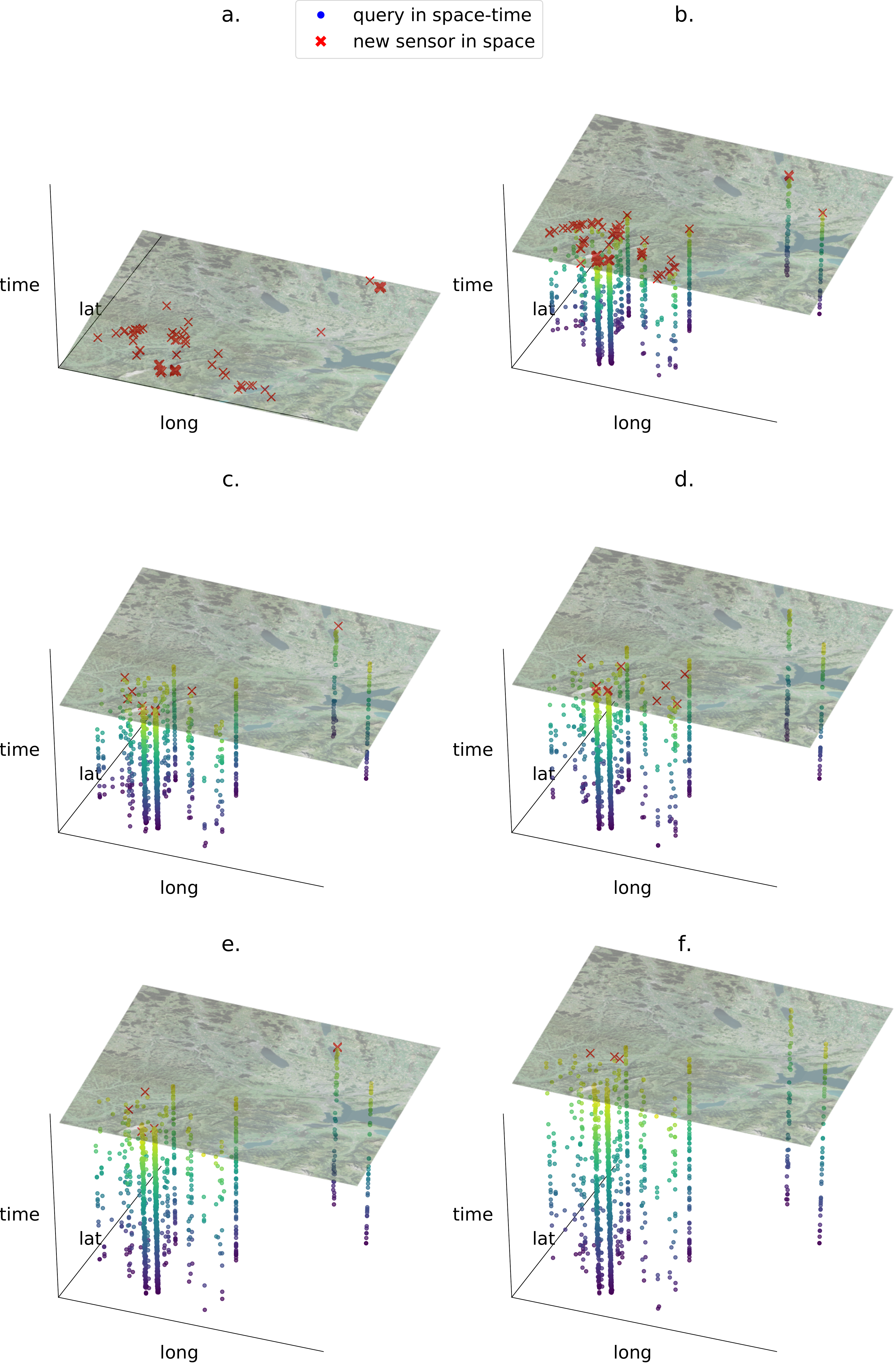}
    \caption{An ADL model shows us where to place sensors and when to collect their data to make better predictions than a PDL model as we proceed in time. We start with choosing a first set of locations for placing sensors uniformly at random (a.). Then, we collect training data for making predictions and learning better feature representations. These show us where to best place new smart meters next (b.). As we collect more data and our feature representations become better, we need fewer new meters to make good predictions (c. - f.).}
\end{figure}

\section*{Methods}

\subsubsection*{The spatio-temporal prediction problem}

Given a map of the Earth, we want to predict some value of interest $\mathbf{y}_{t,\mathbf{s}} \in \mathbb{R}^{D_{y}}$ of dimension $D_{y} \in \mathbb{Z}^{+}$ in time $t \in \mathbb{N}$ and space $\mathbf{s} \in \mathbb{R}^2$ such that $\mathbf{s} = (lat, long)$, with $lat \in [-90, 90]$ and $long \in [-180, 180]$. The ranges of the variables $lat$ and $long$ refer to the possible values of geographic latitudinal and longitudinal coordinates. Hereby, the starting point in time, and the accuracy in both time and space are application dependent and can be chosen arbitrarily. The set of all $\mathbf{y}_{t,\mathbf{s}}$, hereafter called labels, is referred to as $\mathcal{Y}$. Each label is hence a vector

\begin{equation} \nonumber
    \mathbf{y}_{t,\mathbf{s}}  = \begin{pmatrix} y_{t,\mathbf{s}, 1} \\ \vdots \\ y_{t,\mathbf{s}, D_{y}} \end{pmatrix}
\end{equation}

Given the features $\mathbf{x}_{t,\mathbf{s}} \in \mathbb{R}^{D_{x}}$ of dimension $D_{x} \in \mathbb{Z}^{+}$ for each label, we want to predict labels for particular points of interest in time and space. We refer to the set of all features as $\mathcal{X}$. Each label hence  has a corresponding feature vector

\begin{equation} \nonumber
    \mathbf{x}_{t,\mathbf{s}}  = \begin{pmatrix} x_{t,\mathbf{s}, 1} \\ \vdots \\ x_{t,\mathbf{s}, D_{x}} \end{pmatrix}
\end{equation}

We can further classify the single entries of $\mathbf{x}_{t,\mathbf{s}}$ as space, time and space-time variant features. Features that are constant in time $t$ but variant in space $\mathbf{s}$ are referred to as space variant features $\mathbf{x}_{\mathbf{s}} \in \mathbb{R}^{D_s}$ of dimension $D_s \in \mathbb{N}$ such that $D_s \leq D_{x}$:

\begin{equation} \nonumber
    \mathbf{x}_{\mathbf{s}} = \begin{pmatrix} x_{t,\mathbf{s}, 1} \\ \vdots \\ x_{t,\mathbf{s}, D_s} \end{pmatrix} = \begin{pmatrix} x_{t+\hat{t},\mathbf{s}, 1} \\ \vdots \\ x_{t+\hat{t},\mathbf{s}, D_s} \end{pmatrix} \\
    \forall \hat{t} \in \mathbb{Z}: 0 \leq t+\hat{t}
\end{equation}

Features that are constant in space $\mathbf{s}$ but variant in time $t$ are referred to as time variant features $\mathbf{x}_t \in \mathbb{R}^{D_t}$ of dimension $D_t \in \mathbb{N}$ such that $D_t \leq D_{x}$:

\begin{equation} \nonumber
    \begin{split}
        &\mathbf{x}_t = \begin{pmatrix} x_{t, \mathbf{s}, D_s + 1} \\ \vdots \\ x_{t, \mathbf{s}, D_s + D_t} \end{pmatrix} =  \begin{pmatrix} x_{t, \mathbf{s}+\hat{\mathbf{s}}, D_s + 1} \\ \vdots \\ x_{t, \mathbf{s}+\hat{\mathbf{s}}, D_s + D_t} \end{pmatrix} \\
        &\forall \mathbf{\hat{s}} \in \mathbb{R}^2: lat+\hat{lat} \in [-90, 90] \land long+\hat{long} \in [-180, 180]
    \end{split}
\end{equation}

Features that are variant in both time $t$ and space $\mathbf{s}$ are referred to as space-time variant features $\mathbf{x}_{\mathbf{s}t} \in \mathbb{R}^{D_{st}}$ of dimension $D_{st} \in \mathbb{N}$ such that $D_{st} \leq D_{x}$:

\begin{equation} \nonumber
    \begin{split}
        &\mathbf{x}_{\mathbf{s}t} = \begin{pmatrix} x_{t, \mathbf{s}, D_s + D_t + 1} \\ \vdots \\ x_{t, \mathbf{s}, D_s + D_t + D_{st}} \end{pmatrix} \neq  \begin{pmatrix} x_{t+\hat{t}, \mathbf{s}+\hat{\mathbf{s}}, D_s + D_t + 1} \\ \vdots \\ x_{t+\hat{t}, \mathbf{s}+\hat{\mathbf{s}},  D_s + D_t + D_{st}} \end{pmatrix} \\
        &\forall \hat{t} \in \mathbb{Z}, \mathbf{\hat{s}} \in \mathbb{R}^2: 0 \leq t+\hat{t} \land lat+\hat{lat} \in [-90, 90] \land long+\hat{long} \in [-180, 180]
    \end{split}
\end{equation}

For the above to be valid, it further has to hold that

\begin{equation} \nonumber
    D_s +D_t + D_{st} = D_{x}
\end{equation}

and

\begin{equation} \nonumber
    \mathbf{x}_{t,\mathbf{s}}  = \begin{pmatrix} \mathbf{x}_{t}  \\ \mathbf{x}_{\mathbf{s}}  \\ \mathbf{x}_{\mathbf{s}t} \end{pmatrix}
\end{equation}

We can further distinguish the type of predictions that we make in the same fashion. Let $\mathcal{Y}^{avail} \subset \mathcal{Y}$ be a subset of all labels that are available to us, i.e. the ground truth values that we have already measured. Then, for a given point $\mathbf{s}$ in space, if we use any knowledge about $\mathbf{y}_{t_{1},\mathbf{s}}$ such that $t_{1}$ is represented in elements of $\mathcal{Y}^{avail}$ to make predictions about any $\mathbf{y}_{t_{2},\mathbf{s}}$ such that $t_{2}$ is not represented by elements of $\mathcal{Y}^{avail}$, we call this a temporal prediction. Similarly, for a given point $t$ in time, if we use any knowledge about $\mathbf{y}_{t,\mathbf{s}_{1}}$ such that $\mathbf{s}_{1}$ is represented by elements of $\mathcal{Y}^{avail}$ to make predictions about any $\mathbf{y}_{t,\mathbf{s}_{2}}$ such that $\mathbf{s}_{2}$ is not represented by elements of $\mathcal{Y}^{avail}$, we call this a spatial prediction. In contrast, if we make predictions about any $\mathbf{y}_{t,\mathbf{s}}$ such that $(t, \mathbf{s})$ is not represented by elements of $\mathcal{Y}^{avail}$, we call this a spatio-temporal prediction. 

Furthermore, let $\mathcal{X}^{avail}$ be the set of features that are complement to each element of $\mathcal{Y}^{avail}$. Then, we let $\mathcal{D} = (\mathcal{X},\mathcal{Y})$ be the dataset that consists of all feature-label pairs that exist and let $\mathcal{D}^{avail} = (\mathcal{X}^{avail},\mathcal{Y}^{avail}) \subset \mathcal{D}$ be a subset that is available to us.

\subsubsection*{Embedding networks}

Assuming that our dataset $\mathcal{D}^{avail}$ is representative for all values of interest $\mathbf{y}_{t,\mathbf{s}}$, i.e. that samples from $\mathcal{D}^{avail}$ are identically and independently distributed (iid) with some probability distribution $P(\mathbf{x}_{t,\mathbf{s}}, \mathbf{y}_{t,\mathbf{s}})$, allows us to learn a functional relationship $f: \mathcal{X} \rightarrow \mathcal{Y}$ using gradient descent algorithms. Our goal is then to generalize as well as possible on data points that are not in $\mathcal{D}^{avail}$, i.e. we want to perform well on spatial, temporal and spatio-temporal prediction tasks. In practice, however, this is often an invalid assumption as our data distribution $P(\mathbf{x}_{t,\mathbf{s}}, \mathbf{y}_{t,\mathbf{s}})$ can be variant across both space and time. To use the prediction power of DL models and the efficiency of stochastic gradient descent algorithms for training these, we have to design a learning method that is able to tackle such distribution shifts. Here, we consider modular neural network prediction models. For each individual application and its feature types, a large variety of architectures can be sensible. The architectures that we consider require a multi-input structure with at least one separate input for each feature type $\mathbf{x}_{t}$, $\mathbf{x}_{\mathbf{s}}$ and $\mathbf{x}_{\mathbf{s}t}$, and at least one being available for the given prediction task.

Figure 2 shows a general neural network architecture of this type that we introduce as an \textit{embedding network}. Modules of our embedding network can be used to shape encoders that embed features into a vector space of arbitrary dimension. The last layer of each encoder is referred to as an embedding layer. Let $N_{(e)} \in \mathbb{Z}^{+}$ with $(e) \in \{t, s, st, x\}$ be the dimension of the vectors into which $\mathbf{x}_{t,\mathbf{s}}$ can be embedded, i.e. the number of nodes of each embedding layer, and $\mathcal{U}^{(i)}$ be the sets of all embedded vectors in their respective vector spaces with $(i) \in \{time, space, space{\text -}time, joint\}$. With the proposed network architecture, we can then define the prediction model and its feature encoders as functions

\begin{equation} \nonumber
    f_{NN}:\mathcal{X} \rightarrow \mathcal{Y}
\end{equation}

\begin{equation} \nonumber
    enc_{(i)}:\mathcal{X} \rightarrow \mathcal{U}^{(i)}
\end{equation}

\subsubsection*{The data selection problem}

Given $\mathcal{D}^{avail}$ there usually exists a much larger set of data points that are not available to us. We refer to these as the candidate features $\mathcal{X}^{cand}$ and their complement labels $\mathcal{Y}^{cand}$ which together shape the set of candidate data points $\mathcal{D}^{cand} = (\mathcal{X}^{cand},\mathcal{Y}^{cand})$. Our goal is to choose the most informative subset of labels $\mathcal{Y}^{choice(*)}$ from the large pool of candidate labels $\mathcal{Y}^{cand}$ such that our overall generalization error decreases the most, without exceeding a given number of labels, which we refer to as our data budget $n_{budget}$. During the data selection process, we assume to have complete access to all existing features $\mathcal{X}$, the available labels $\mathcal{Y}^{avail}$, but not to any labels from $\mathcal{Y}^{cand}$. It hence has to hold that

\begin{equation} \nonumber
    \mathcal{D}^{avail} \cup \mathcal{D}^{cand} = \mathcal{D}
\end{equation}

\begin{equation} \nonumber
    \mathcal{D}^{avail} \cap \mathcal{D}^{cand} = \varnothing
\end{equation}

\begin{equation} \nonumber
    \mathcal{Y}^{choice(*)} \subset \mathcal{Y}^{cand}
\end{equation}

The subset of labels that we eventually query without prior information about their values is likely to deviate from the optimal subset $\mathcal{Y}^{choice(*)}$; we refer to the actually queried subset of labels with $\mathcal{Y}^{choice}$. The feature-label pairs of queried data points are respectively referred to as $\mathcal{D}^{choice} = (\mathcal{X}^{choice}, \mathcal{Y}^{choice})$. One way to query labels is to do so one by one. Another, computationally more efficient way to do this, is to use batches of data queries, particularly because we also train our neural network models with batches of data points between each ADL iteration. We define the batch size, or number of labels, that are queried in each step of an ADL process as $n_{batch} \in \mathbb{Z}^{+}$ and the total number of data selection steps as $n_{iter} \in \mathbb{Z}^{+}$. It hence has to hold that

\begin{equation} \nonumber
    n_{iter} \cdot n_{batch} \leq n_{budget}
\end{equation}

\subsubsection*{Embedded feature vectors}

Given any of the encoders $enc_{(i)}$ that $f_{NN}$ incorporates, we can encode each feature vector $\mathbf{x}_{t,\mathbf{s}}$, and single parts of it ($\mathbf{x}_{t}$, $\mathbf{x}_{\mathbf{s}}$ or $\mathbf{x}_{\mathbf{s}t}$), into their embedded vector spaces. We expect the distances of these vectors to each other to become increasingly meaningful in the context of our overarching prediction task as we train the actual prediction model $f_{NN}$ \cite{Mikolov2013, Frome2013, Pennington2014, Perozzi2014, Devlin2019}. As our encoders are modules of our prediction model, they are automatically trained each time we apply backpropagation on $f_{NN}$ through stochastic gradient descent. Mutual information between parts of our feature vector can propagate back into each encoder, such that encoded parts of our feature vector can preserve information about the all features and our method becomes sparse. For every feature vector $\mathbf{x}_{t,\mathbf{s}}$ and $(i) \in \{time, space, space{\text -}time, joint\}$, we can write

\begin{equation} \nonumber
    \forall \mathbf{x}_{t,\mathbf{s}}  = \begin{pmatrix} \mathbf{x}_{t}  \\ \mathbf{x}_{\mathbf{s}}  \\ \mathbf{x}_{\mathbf{s}t} \end{pmatrix} \in \mathcal{X} \exists \{\hat{\mathbf{x}}_{(j)} = enc_{(i)}(\mathbf{x}_{(j)})\}_{(j) = \{(t,\mathbf{s}), t, \mathbf{s}, \mathbf{s}t\}}
\end{equation}

We refer to predicted labels of data points as

\begin{equation} \nonumber
    \hat{\mathbf{y}}_{t,\mathbf{s}} = f_{NN}(\mathbf{x}_{t,\mathbf{s}})
\end{equation}

\subsubsection*{Clusters of embedded feature vectors}

Given a set of vectors of the same dimension, we can calculate clusters based on the distances of these vectors to each other using algorithms like K-means or affinity propagation. In order to execute most clustering algorithms, we need to determine the number of desired clusters or a minimum distance of members beforehand. In order to avoid assumptions regarding common distances in the embedded vector spaces, we only consider clustering methods that require a definition of the number of clusters beforehand. We refer to the number of clusters that we set for performing any of these clustering methods with $n_{clusters}$. Let $n_{diverse}^{(i)}$ further be the number of distinct elements in the vector set $\mathcal{U}^{(i)}$. For a clustering of embedded vectors to be valid it hence has to hold that

\begin{equation} \nonumber
    n_{clusters} < n_{diverse}^{(i)}
\end{equation}

and for data queries to be sensible furthermore that

\begin{equation} \nonumber
    n_{clusters} << n_{diverse}^{(i)}
\end{equation}

After clustering the elements of any embedded vector set $\mathcal{U}^{(i)}$ with $(i) \in \{time, space, space{\text -}time, joint\}$, we get a first set of vectors $\mathbf{c}^{(i)}_{l}$ which describe the center of each cluster with $l=1...n_{clusters}$, and a set of values $m^{(i)}_{k}$ which describe the membership IDs for each clustered data point with
$k=1...|\mathcal{U}^{(i)}|$.

\subsubsection*{Embedding uncertainty: distance of features to cluster centers}

The distance between any two vectors of the same dimension can be calculated using inner products through e.g. kernel functions or other distance measures such as the cosine similarity. Using any of these measures, we can calculate the distance $d_{c, (j)}$ of every embedded vector $\hat{\mathbf{x}}_{(j)}$ to its corresponding cluster center $\mathbf{c}^{(i)}_{m^{(i)}_{k}}$ with  $(i) \in \{time, space, space{\text -}time, joint\}$, $(j) \in \{t, \mathbf{s}, \mathbf{s}t, (t, \mathbf{s})\}$, and $k = 1,...,|\mathcal{U}^{(i)}|$ being the element ID that corresponds to the point $(t, \mathbf{s})$. We use this distance as a metric of uncertainty, called the \textit{embedding uncertainty} of our ADL method. Alternatively, we can also use a Gaussian mixture model to cluster embedded feature vectors and express uncertainties in a single step.

\subsubsection*{Batch active learning algorithm}

Given our embedding network DL model and our embedding uncertainty metric of informativeness $d_{c, (j)}$, we create a pool-based ADL method that queries a batch of labels $\mathbf{y}_{t,\mathbf{s}} \in \mathcal{Y}^{cand}$ from the candidate data pool in each iteration. Algorithm 1 provides a pseudo-code for this, which we go through in detail.

Starting with a DL model $f^{NN}$ that is trained on randomly chosen initial data (1.), we can choose which feature type and corresponding encoder we want to use for querying candidate data points (2.). Given some data budget $n_{budget}$ and a maximum number of iterations $n_{iter}$, we create a data counter $c_{budget}$ and an iteration counter $c_{iter}$ that we set to zero, and leave the set of queried data points empty prior to performing ADL (3.). We start our ADL iterations by encoding each candidate data point (4.1). If the set of candidate data points is too large for this to be computationally feasible, we can sample a subset of candidate data points uniformly at random. Next, we cluster all embedded feature vectors (4.2) and compute the distances to their respective cluster centers (4.3). We can then pick the most informative data point from each cluster using one of our ADL variants (4.4). The chosen subset of data points is then used for training our prediction model $f_{NN}$ (4.5). Here, arbitrary techniques like weight regularization, early stopping and adaptive learning rates can be used to enhance training. Next, we remove queried data points from the candidate data pool at a rate $\delta \in \mathbb{R}$ with $0 \leq \delta \leq 1$ (4.6). Here, the rate $\delta$ is the probability with which we remove a queried data point. A value of $\delta=1$ means that all queried data points are removed and a value of $\delta=0$ respectively means that all queried data points are kept. Before we continue with the next iteration, we update the set of queried candidates (4.7), increment our data point counter by the number of newly queried labels among the chosen data points (4.8) and our iteration counter by one (4.9).  

We can highlight two major differences to existing ADL methods. First, we remove queried data points from the candidate dataset at some rate $\delta$; this allows us to re-use data points so as to implicitly reduce ($\delta \rightarrow 0$) or increase ($\delta \rightarrow 1$) a bias towards yet unqueried data points. Second, we set the number of clusters of our candidates equal to the batch size of data points that we want to query in each iteration; this allows us to implicitly sample more points by building more clusters where data points are densely populated, hence having a sufficiently representative sample of our entire data population. Simultaneously, this allows us to cope with imbalanced data as we also create clusters where data is located in isolation and available in small amounts in the encoded space. 

Memory and time complexity of the algorithm depend on the size of our candidate data pool $|\mathcal{D}^{cand}|$ and can both be reduced by down-sampling the pool at the cost of less optimal data queries. The time complexity of our algorithm further depends on the method we use to cluster embedded feature vectors (4.2), the method we use to compute the distance of candidate data to cluster centers (4.3), and the complexity of training our DL model (4.5). For most prediction tasks, it is realistic to assume that the complexity of (4.5) is smaller than (4.2) and (4.3). If we for instance use the K-means++ algorithm to cluster embedded feature vectors and calculate the Manhattan distance between embedded vectors and their cluster center, it holds that

\begin{equation} \nonumber
    O(|\mathcal{D}^{cand}| \cdot N_{(e)} \cdot log( N_{(e)} ) (4.3) <
    O(|\mathcal{D}^{cand}| \cdot N_{(e)} \cdot n^{batch}) (4.2)
\end{equation}

and the computational complexity of our algorithm breaks down to that of our clustering algorithm (4.2), if we further make the realistic assumption that 

\begin{equation} \nonumber
    log(N_{(e)}) << n^{batch}
\end{equation}

\begin{algorithm}[H]
 1. Train $f_{NN}$ on $D^{avail}$. \\
 2. Choose $(j) \in \{t, \mathbf{s}, \mathbf{s}t, (t,\mathbf{s})\} \leftrightarrow (i) \in \{time, space, space{\text -}time, joint\}$. \\
 3. $c_{budget} = 0; c_{iter} = 0; \mathcal{D}^{choice} = \varnothing$. \\
 \While{$c_{budget} < n_{budget}$ and $c_{iter} \leq n_{iter}$}{
  
  \If{$|\mathcal{X}^{cand}| >> n_{budget}$}{
   $\mathcal{X}^{cand} \leftarrow \mathcal{X}^{subset} \subset \mathcal{X}^{cand}$
  }
  4.1 $\hat{\mathbf{x}}_{(j)} = enc_{(i)}(\mathbf{x}_{(j)}) \forall \mathbf{x}_{t,\mathbf{s}} \in \mathcal{X}^{cand}$\\
  4.2 Cluster $\hat{\mathbf{x}}_{(j)}$ with $n_{clusters} = n_{batch}$. \\
  4.3 Compute $d_{c, (j)}$. \\
  4.4 $\mathcal{Y}^{choice}_{(c_{iter})} = \{\hat{\mathbf{x}}_{(j)}\}_{k=1}^{n_{batch}}$.\\
  4.5 Train $f_{NN}$ on $\mathcal{D}^{choice}_{(c_{iter})} = (\mathcal{X}^{choice}_{(c_{iter})}, \mathcal{Y}^{choice}_{(c_{iter})})$. \\
  4.6 $ \mathcal{D}^{cand} \leftarrow \mathcal{D}^{cand} \setminus \mathcal{D}^{choice}_{(c_{iter})}$ at rate $\delta$ \\
  4.7 $\mathcal{D}^{choice} \leftarrow \mathcal{D}^{choice} \cup \mathcal{D}^{choice}_{(c_{iter})}$ \\
  4.8 $c_{budget} \leftarrow |\mathcal{D}^{choice}|$ \\
  4.9 $c_{iter} \leftarrow c_{iter} + 1$ \\
 }
 \caption{A pseudo-code of the proposed batch ADL method.}
\end{algorithm}

\subsubsection*{Datasets}

We are given the electric consumption measurements of 100-400 buildings in Switzerland in 15-min steps from local distribution system operators. Using the geographic coordinates of these buildings, we further collect aerial imagery of each building with a resolution of 25 cm per pixel \cite{GeoVITE2020}. We then cluster all buildings that are in a distance of at most 1 km to each other. For each cluster of buildings, we calculate the cluster centers and collect a total of nine meteorological time-series measurements from reanalysis data for each of these clusters with one hour accuracy \cite{Pfenninger2016, Staffel2016}. The meteorological values that we use consist of air density in kg/m³, cloud cover, precipitation in mm/hour, ground level solar irradiance in W/m², top of atmosphere solar irradiance W/m², air temperature in °C, snowfall in mm/hour, snow mass in kg/m² and wind speed.

We predict the next 24 hours of electric consumption. For each of the nine meteorological conditions, we consider a historic time window of 24 hours. Time stamps are ordinal encoded, and contain information about the month, day, hour and quarter-hour in which the corresponding electric consumption of a building occurs. Images of buildings are processed using histograms of their pixel values with 100 bins for each image channel (red, green, blue). We can hence set the dimensions of the feature and label vectors to

\begin{equation} \nonumber
    D_{t} = 4
\end{equation}

\begin{equation} \nonumber
    D_{s} = 3 \cdot 100 = 300
\end{equation}

\begin{equation} \nonumber
    D_{st} = 9 \cdot 24 = 216
\end{equation}

\begin{equation} \nonumber
    D_{x} = D_{t} + D_{s} + D_{st} = 520 
\end{equation}

\begin{equation} \nonumber
    D_y = 4\cdot 24 = 96
\end{equation}

\subsubsection*{Training, validation and testing data}

Given a number of data points that are available to us, we create training, validation and testing data for our hypothesis test. The training data is used to fit our prediction model prior to performing ADL. The validation data is used to avoid that we overfit our model to the training data through early stopping. The testing data is used as the candidate data pool on which we perform ADL to train our prediction model. We separate our testing data into spatial, temporal and spatio-temporal prediction tests. We use $3\%$ of our data for initial training, $6\%$ for validation, and $91\%$ for testing. We further split our testing data such that $23\%$ of it represent spatial predictions, another $23\%$ temporal predictions and $54\%$ spatio-temporal predictions. In the following, we will refer to training, validation and testing data with $\mathcal{D}^{train}, \mathcal{D}^{val}$ and $\mathcal{D}^{test}$. We let $\mathcal{D}$ be our entire feature-label space and can write for initially available and candidate data that

\begin{equation} \nonumber
    \frac{|\mathcal{D}^{avail}|}{|\mathcal{D}|} = 0.03
\end{equation}

\begin{equation} \nonumber
    \frac{|\mathcal{D}^{cand}|}{|\mathcal{D}|} = 0.91
\end{equation}

\subsubsection*{Embedding network prediction model and feature encoders}

We construct our DL model from multiple subnetworks. The network which processes meteorological data consists of a 1-D convolutional neural network layer with 16 filters. The networks which process time stamp data and the histograms of building image pixels each contain one densely connected hidden layer with 1,000 nodes. The joint encoder concatenates the outputs of each of these networks and adds another densely connected hidden layer with 1,000 nodes. All embedding layers consist of 100 nodes. The entire prediction model then takes the output of the joint encoder and adds another layer of 1,000 densely connected nodes before mapping the joint inputs to the desired output with 96 densely connected nodes. In total, our model contains 10,744,600 trainable and zero non-trainable parameters. For the encoder outputs of all $(e) \in \{t, s, st, x\}$,  we can write

\begin{equation} \nonumber
    N_{(e)} = 100
\end{equation}

\subsubsection*{Loss function}

We use the MSE, also known as the $L2$ loss, between predicted labels $\hat{\mathbf{y}}_{t,\mathbf{s}}$ and true labels $\mathbf{y}_{t,\mathbf{s}}$ as our loss $L(\hat{\mathbf{y}}_{t,\mathbf{s}}, \mathbf{y}_{t,\mathbf{s}})$. One can equivalently use the mean absolute error (MAE), also known as the $L1$ loss, or variations from these with minor impacts on our empirical tests. In each epoch of training and validation, as well as for each test, we calculate the total loss function $loss(\mathcal{D}^{(d)})$ as the average loss of all data points in the respective datasets $D^{(d)}$, where $(d) \in \{train, val, test\}$. With $j = 1,...,|\mathcal{D}^{(d)}|$ being the $j$-th element that corresponds to the point $(t, \mathbf{s})$ in $\mathcal{D}^{(d)}$, we can write for all pairs of $(\hat{\mathbf{y}}_{j}, \mathbf{y}_{j}) \in \mathcal{D}^{(d)}$ that

\begin{equation} \nonumber
    L(\hat{\mathbf{y}}_{t,\mathbf{s}}, \mathbf{y}_{t,\mathbf{s}}) = \frac{\sum_{k=1}^{D_y} (y_{t,\mathbf{s}, k}-\hat{y}_{t,\mathbf{s}, k})^2}{D_y}
\end{equation}

and 

\begin{equation} \nonumber
    loss(\mathcal{D}^{(d)}) = \frac{\sum_{j=1}^{|\mathcal{D}^{(d)}|} L(\hat{\mathbf{y}}_{j}, \mathbf{y}_{j})}{|\mathcal{D}^{(d)}|}
\end{equation}

\subsubsection*{Experiments}

We assume our data budget to be $50\%$ of the size of our candidate data pool, i.e. we want to choose the more informative half of candidate data points. We perform ten iterations of the above Algorithm 1 where we query 10\% of our data budget in each iteration. We train our prediction model for 30 epochs and use an early stopping patience of 10 epochs when training our prediction model on the initially available data and in each iteration of Algorithm 1. We can write

\begin{equation} \nonumber
    n_{budget} = 0.5 \cdot |\mathcal{D}^{cand}|
\end{equation}

\begin{equation} \nonumber
    n_{iter} = 10
\end{equation}

\begin{equation} \nonumber
    n_{clusters} = n_{batch} = 0.1 \cdot n_{budget} = 0.05 \cdot |\mathcal{D}^{cand}|
\end{equation}

We use the K-means++ algorithm to cluster embedded feature vectors, and the Laplacian kernel to calculate the distance between each embedded feature vector and its cluster center. Given the embedded vector set $\mathcal{U}^{(i)}$ with $(i) \in \{time, space, space\text{-}time, joint\}$, the vectors that describe the center of each cluster $\mathbf{c}^{(i)}_{l}$ with $l=1...n_{clusters}$ and a membership ID $m^{(i)}_{k}$ for each embedded feature $k=1...|\mathcal{U}^{(i)}|$, we calculate the distance $d_{c, (j)}$ for every feature type $(j)\in \{t, \mathbf{s}, \mathbf{s}t, (t,\mathbf{s})\}$, corresponding encoder outputs $(e)\in \{t, s, st, x\}$ and point in time-space $(t, \mathbf{s})$ as

\begin{equation} \nonumber
    d_{c, (j)} = exp(- \frac{|| \hat{\mathbf{x}}_{(j)} - \mathbf{c}^{(i)}_{m^{(i)}_{k}} ||_1}{N_{(e)} })
\end{equation}

We test Algorithm 1 for every partial feature vector, and the entire feature vector separately. Since our labels have a similar dimension ($D_y = 96$) as our embedded features ($N_{(e)} = 100$), we use the predicted labels ($\hat{\mathbf{y}}_{t,\mathbf{s}}$) as our jointly embedded feature vectors ($\hat{\mathbf{x}}_{t,\mathbf{s}}$). We can also design our joint feature encoder to contain all layers of our DL prediction model except for the output layer. This is proposed in \cite{Ash2020} and represents a special case of the method we propose. We also evaluate a scenario in which we query candidate data points based on the distance of their true labels $\mathbf{y}_{t,\mathbf{s}}$. We conduct our tests for the two cases that we mainly distinguish: first, we remove each queried point from the candidate data pool at the end of each ADL iteration ($\delta=1$); second, we keep queried data points in the candidate pool throughout all ADL iterations ($\delta=0$).

\section*{Data availability}

The meteorological data that we use in our experiments were made available from the renewables.ninja platform \cite{RenewablesNinja2020}. In order to maintain the privacy of our data providers, the electric consumption data and aerial imagery of buildings that we use in our empirical tests cannot be made available. Instead, we provide a toy dataset that consists of sampled load profiles and histograms of pixel values of building images which can be used to reproduce all elements of our original experiments in a Github repository \cite{Aryandoust2021}. Data on global smart meter adoption is available under 'DataSelectionMaps/images/global meter deployment'.

\section*{Code availability}

We provide step-by-step instructions for implementing the algorithm that we apply in a Jupyter notebook session on Github \cite{Aryandoust2021}. We further maintain a Python package \cite{Aryandoust2021_2} and Docker container image \cite{Aryandoust2021_3} for performing ADL with our proposed algorithm on arbitrary spatio-temporal prediction tasks, all made available under an MIT license.

\section*{Acknowledgements}
We acknowledge funding from the European Union's Horizon 2020 research and innovation programme under grant agreement No 837089, for the SENTINEL project.

\section*{Author contributions}

Arsam Aryandoust conceptualised the research, designed and implemented all methods, conducted all analyses, and drafted the manuscript. Stefan Pfenninger edited and revised the manuscript and reviewed the code. Anthony Patt helped in analysing the data and supervised the creation of the manuscript.

\section*{Competing interests}

The authors declare no competing financial or non-financial interests.


\newpage
\section*{Appendix}

\subsection*{Appendix 1 - Artificial intelligence, machine-, deep-, passive- and active-learning}

Artificial intelligence (AI) is the general study of leveraging computation and data for solving real-world problems that humanity faces. A subfield of AI that can provide predictions for solving many of these real-world problems is machine learning (ML). ML allows us to solve a variety of problems by either learning functional relationships between labelled data using supervised methods, or by learning patterns in unlabelled data using unsupervised methods. In this, deep learning (DL) has become a particularly important subfield of ML. DL includes the use of artificial neural networks, which are prediction models that have proven to be particularly useful because they are able to learn complex function classes for representing patterns and relationships in data that other machine learning models like for instance random forests, support vector machines or linear regression models are not capable of \cite{LeCun2015}. DL prediction models can be trained using either a passive learning approach, which is today's default method of choice, or using an active learning approach, which is emerging more recently.

In passive learning, the data that is used for training a DL prediction model is collected from a large pool of candidate data points at random. This is useful in domains where data can be collected at zero to very low costs. An example for such a domain are repeatedly running Atari games, in which an agent chooses random actions in every new state of the game in order to explore the expected outcome of the game in response to its action choices \cite{Schrittwieser2020}. Here, the set of state-action pairs are features and the corresponding outcomes of the game are labels that together form the data points that are randomly collected using passive learning. In many other domains, however, randomly collecting data naturally results in a large demand for data and, wherever the spatial distribution of sensors is involved, a sub-optimal deployment of sensors. This makes DL with a passive learning approach inapplicable to many real-world problems where the budget for deploying sensors and labelling data is limited. Predicting electric load in time and space is one of these tasks: we are limited in the number of smart metering devices that we can deploy in space for measuring the electric load at every building. Further, we are also limited in the temporal granularity of the data that we measure because too fine-grained measurements, say more than one measurement of weather or consumption per minute from each smart meter or weather station, would result in massive amounts of unprocessable data \cite{Hejazi2018}. Another example for such a domain is water contamination detection, where only a small number of sensors must be optimally deployed over time and space to monitor and predict the pollution of a large water network in the early stage of an incident \cite{Shahra2020}. Hence, a fundamental question that arises for predicting electric load and many other prediction tasks is where and when to measure and query the data required to make the best possible predictions while staying within a maximum budget for sensors and data \cite{Hong2015}.

In active learning, as opposed to passive learning, we leverage additional computation to assess the informativeness of data points to collect only the most informative subset of data from a large pool of candidates, and reduce our overall demand for data, or deploy sensors more optimally in time and space. In our task, we are given a small budget for data that we can label and a large pool of candidate data points that are distributed over time and space, and want to iteratively query a batch of data labels which let us train a prediction model that generalizes well on all unqueried data points. In each iteration, this is a pool-based batch active learning problem, as opposed to a stream-based active learning problem where we would sequentially decide on whether to query a data label from a stream of candidate data points. Existing pool-based active learning methods can roughly be categorized into two groups \cite{Settles2010, Dasgupta2011, Kumar2020}: a first group of methods tries to shrink the space of candidate prediction models as fast as possible, while a second group of methods tries to exploit patterns in the underlying data. The first group of methods can further be categorized into uncertainty \cite{Lewis1994_2, Lindenbaum2004, Culotta2005, Koerner2006, Schein2007}, disagreement \cite{Dagan1995, McCallum1998, Muslea2000, Melville2004, Hanneke2014}, variance reduction \cite{MacKay1992, Cohn1996, Zhang2000, Hoi2006}, error reduction \cite{Roy2001, Zhu2003, Guo2007, Moskovitch2007} and model change based \cite{Settles2007, Settles2008} methods. The second group of methods is less diverse and distinguishes mainly density weighted \cite{Settles2007, Settles2008} and clustering based \cite{Nguyen2004, Xu2007, Dasgupta2008} methods. 

\subsection*{Appendix 2 - Choice of feature and label data}
We use the satellite or aerial image of a building as a spatial feature of a building. This is an unconventional approach for predicting electric load profiles. More commonly used data for spatial features of a building are for instance its type of insulation and occupancy. Although these might contain valuable information, leading to a higher prediction accuracy of electric load, they represent ground truth data that is difficult to collect. Using remotely sensed data as features like satellite images and meteorological conditions derived from these, allows us to scale our prediction task, without the need of collecting additional ground truth data. This justifies our assumption that our features are available at very low to zero costs, and we can formulate our data selection task as a pool-based active learning problem. Our DL model will automatically omit unimportant features and extract valuable ones like insulation and size of a building visible in an image in their mutual context with meteorological data and time, and map these to the electric load as our only ground truth data. Different than in spatio-temporal point processes, we further do not condition our predictions on any historic events about the labels of our prediction task \cite{Gonzalez2016}. This means that we do not use any past values of electric load as features for making predictions of load in the future. This allows us to make predictions for any point in time and space, without having ground truth data available for these points beforehand.

\subsection*{Appendix 3 - Alternative active learning methods}

We can apply alternative active learning methods by introducing uncertainties through a probability distribution or a variance over possible predictions. First, we could transform our regression problem into a multi-class classification task where each class represents a certain range of electric consumption. We can then predict classes instead of continuous floating point numbers. This would directly give us uncertainties about our predictions for each candidate data. Second, we could predict parameters of a probability distribution over each consumption value and read our model uncertainties from the concentration of our distributions over all of these possible values. Third, we could use random weight drop-outs during inference to express uncertainties through the variance we get for the prediction of each candidate data point from a single prediction model that we train. Forth, we could design a multi headed neural network that simultaneously makes multiple predictions of the same profiles and use these multiple predictions as ensembles to express variance and uncertainty through the degree of consensus among all outputs. Similarly, we could train multiple prediction models and express uncertainties through distinct ensembles.

The advantage of the active learning method we propose here over existing ones is that it is sparse: given our embedding network, we are able encode single parts of our feature vector only for expressing the informativeness of candidate data points, and decide which data point to query. For example, if the aerial image of a building is expensive to collect or not available, we are able to encode the meteorological conditions in the region of that building to asses the informativeness of candidate data points, and vice versa. A potential disadvantage of applying our method compared to existing ones is that we might achieve smaller increases in prediction accuracy and data saving, as well as have a larger computational overhead for assessing the informativeness of candidate data points. However, comparing our method with existing ones is beyond the scope of this study and is left for a further research opportunity.

\subsection*{Appendix 4 - Results for 100 buildings and ADL variables}

With the ADL variables $\mathbf{x}_{t}$ and $\mathbf{x}_{s}$, we generally make equally good or slightly worse predictions than with PDL. For $\delta=0$, $\mathbf{x}_{t}$ and $\mathbf{x}_{s}$ either use the same amount of data as PDL (rnd $d_{c}$), or an amount of data that is close to the batch size of data points that we query in each iteration of the algorithm (max/min $d_{c}$). For $\delta=1$, all ADL variants for $\mathbf{x}_{t}$ and $\mathbf{x}_{s}$ perform similar to PDL. This is because the distinct points in time $\mathbf{x}_{t}$ and space $\mathbf{x}_{s}$ contained in our data are smaller than the batch size and number of clusters in the data points that we query in each iteration of our ADL algorithm. Therefore our algorithm performs identical to PDL for these variables with $\delta=1$. For $\delta=0$, when we collect candidate data points evenly distributed from all clusters, we also collect data points uniformly at random from the entire candidate data pool and hence proceed equal to PDL. If we instead maximize or minimize the embedding distance of candidate data points, we choose about the same set of data points in each iteration and hence observe a data usage that is close to the batch size of the data that we query in a single iteration (10-15\%), which performs poor when generalizing. Table 2 contains the numerical results for experiments that we run on our second dataset that is composed of profiles from 100 distinct buildings.

\begin{table}
    \caption{Numerical results for each prediction type, ADL variable and ADL variant with experiments on dataset with 100 buildings. The columns named 'data' present how percent of data budget is used; 'sensors' state what percent of sensors are used from the new sensors initially available in the candidate data pool; 'test loss' states the average L2 loss on yet unqueried data; 'accuracy' is calculated as $1 - min(1, \frac{\text{PDL loss}}{\text{RF loss}})$ and $1 - min(1, \frac{\text{ADL loss}}{\text{RF loss}})$.}
    \begin{center}
    \scalebox{0.7}{
        \begin{tabular}{ ccc|c|c|c|c|c|c|c|c| }
         \cline{4-11}
         & & & \multicolumn{4}{ c| }{Removing queried data ($\delta=1$)} & \multicolumn{4}{ c| }{Keeping queried data ($\delta=0$)}\\
         \hline
         \multicolumn{1}{ |c }{prediction type} & ADL variable & ADL variant & data & sensors & accuracy & test loss & data & sensors & accuracy & test loss \\ \hline
          \multicolumn{1}{ |c }{Spatial}& $\mathbf{x}_{st}$ & rnd $d_{c, st}$ & 100\% & 100\% & 37\% & 0.243801 & 47\% & 100\% & 87\% & 0.169622 \\ 
          \multicolumn{1}{ |c }{ }& & max $d_{c, st}$ & 100\% & 100\% & 28\% & 0.279894 & 18\% & 65\% & 28\% &  0.967486 \\ 
          \multicolumn{1}{ |c }{ }& & min $d_{c, st}$ & 100\% & 100\% & 39\% & 0235388 & 20\% & 100\% & 74\% &  0.346473 \\ 
          \multicolumn{1}{ |c }{ }& & avg $d_{c, st}$ & 100\% & 100\% & 32\% & 0.263962 & 18\% & 68\% & 54\% & 0.617858 \\  
          \multicolumn{1}{ |c }{ }& $\mathbf{x}_{t,s}$ & rnd $d_{c, (t,s)}$ & 100\% & 100\% & 72\% & 0.107482 & 75\% & 100\% & 84\% & 0.210018 \\ 
          \multicolumn{1}{ |c }{ }& & max $d_{c, (t,s)}$ & 100\% & 100\% & 74\% & 0.102124 & 57\% & 100\% & 75\% & 0.336121 \\ 
          \multicolumn{1}{ |c }{ }& & min $d_{c, (t,s)}$ & 100\% & 100\% & 78\% & 0.086680 & 72\% & 100\% & 71\% & 0.390072 \\ 
          \multicolumn{1}{ |c }{ }& & avg $d_{c, (t,s)}$ & 100\% & 100\% & 74\% & 0.100421 & 66\% & 100\% & 80\% & 0.265017 \\ 
          \multicolumn{1}{ |c }{ }& $\mathbf{\hat{y}_{t,s}}$ & rnd $d_{c, (t,s)}$ & 100\% & 100\% & 92\% & 0.032150 & 46\% & 100\% & 93\% & 0.099304 \\ 
          \multicolumn{1}{ |c }{ }& & max $d_{c, (t,s)}$ & 100\% & 100\% & 92\% & 0.030691 & 40\% & 95\% & 92\% & 0.104101 \\ 
          \multicolumn{1}{ |c }{ }& & min $d_{c, (t,s)}$ & 100\% & 100\% & 91\% & 0.035534 & 48\% & 100\% & 92\% & 0.104086 \\ 
          \multicolumn{1}{ |c }{ }& & avg $d_{c, (t,s)}$ & 100\% & 100\% & 92\% & 0.031574 & 40\% & 98\% & 92\% & 0.114373 \\ 
          \multicolumn{1}{ |c }{ }& $\mathbf{y}_{t,s}$ & rnd $d_{c, (t,s)}$ & 100\% & 100\% & 91\% & 0.033857 & 39\% & 100\% & 78\% & 0.298844 \\ 
          \multicolumn{1}{ |c }{ }& & max $d_{c, (t,s)}$ & 100\% & 95\% & 91\% & 0.035090 & 21\% & 74\% & 66\% & 0.461655 \\ 
          \multicolumn{1}{ |c }{ }& & min $d_{c, (t,s)}$ & 100\% & 98\% & 90\% & 0.038516 & 24\% & 83\% & 72\% & 0.379143 \\ 
          \multicolumn{1}{ |c }{ }& & avg $d_{c, (t,s)}$ & 100\% & 97\% & 92\% & 0.030082 & 23\% & 82\% & 69\% & 0.414389 \\  \hline
          \multicolumn{3}{ |c| }{passive deep learning (PDL) benchmark} & 100\% & 100\% & 32\% & 0.261937 & 80\% & 100\% & 46\% & 0.729953 \\  \hline
          \multicolumn{3}{ |c| }{random forest (RF) baseline} & 0\% & 0\% & 0\% & 0.387128 & 0\% & 0\% & 0\% & 1.352659 \\  \hline
          \multicolumn{1}{ |c }{Temporal}& $\mathbf{x}_{st}$ & rnd $d_{c, st}$ & 100\% & 0\% & 69\% & 0.439798 & 75\% & 0\% & 35\% & 0.198148 \\ 
          \multicolumn{1}{ |c }{ }& & max $d_{c, st}$ & 100\% & 0\% & 65\% & 0.497277 & 15\% & 0\% & 0\% & 0.354400 \\ 
          \multicolumn{1}{ |c }{ }& & min $d_{c, st}$ & 100\% & 0\% & 61\% & 0.550875 & 19\% & 0\% & 15\% & 0.257450 \\ 
          \multicolumn{1}{ |c }{ }& & avg $d_{c, st}$ & 100\% & 0\% & 67\% & 0.463081 & 15\% & 0\% & 0\% & 0.327369 \\ 
          \multicolumn{1}{ |c }{ }& $\mathbf{x}_{t,s}$ & rnd $d_{c, (t,s)}$ & 100\% & 0\% & 69\% & 0.432739 & 74\% & 0\% & 56\% & 0.134869 \\ 
          \multicolumn{1}{ |c }{ }& & max $d_{c, (t,s)}$ & 100\% & 0\% & 72\% & 0.396843 & 52\% & 0\% & 31\% & 0.0209475 \\ 
          \multicolumn{1}{ |c }{ }& & min $d_{c, (t,s)}$ & 100\% & 0\% & 73\% & 0.382284 & 70\% & 0\% & 41\% & 0.180766 \\ 
          \multicolumn{1}{ |c }{ }& & avg $d_{c, (t,s)}$ & 100\% & 0\% & 69\% & 0.441338 & 61\% & 0\% & 42\% & 0.176255 \\ 
          \multicolumn{1}{ |c }{ }& $\mathbf{\hat{y}}_{t,s}$ & rnd $d_{c, (t,s)}$ & 100\% & 0\% & 94\% & 0.081201 & 47\% & 0\% & 67\% & 0.101231 \\ 
          \multicolumn{1}{ |c }{ }& & max $d_{c, (t,s)}$ & 100\% & 0\% & 94\% & 0.078725 & 35\% & 0\% & 61\% & 0.117624 \\ 
          \multicolumn{1}{ |c }{ }& & min $d_{c, (t,s)}$ & 100\% & 0\% & 94\% & 0.087990 & 47\% & 0\% & 64\% & 0.110295 \\ 
          \multicolumn{1}{ |c }{ }& & avg $d_{c, (t,s)}$ & 100\% & 0\% & 94\% & 0.080757 & 37\% & 0\% & 63\% & 0.113722 \\ 
          \multicolumn{1}{ |c }{ }& $\mathbf{y}_{t,s}$ & rnd $d_{c, (t,s)}$ & 100\% & 0\% & 94\% & 0.086061 & 43\% & 0\% & 63\% & 0.112570 \\ 
          \multicolumn{1}{ |c }{ }& & max $d_{c, (t,s)}$ & 100\% & 0\% & 93\% & 0.093925 & 24\% & 0\% & 0\% & 0.324053 \\ 
          \multicolumn{1}{ |c }{ }& & min $d_{c, (t,s)}$ & 100\% & 0\% & 94\% & 0.082089 & 27\% & 0\% & 46\% & 0.163093 \\ 
          \multicolumn{1}{ |c }{ }& & avg $d_{c, (t,s)}$ & 100\% & 0\% & 94\% & 0.082410 & 26\% & 0\% & 20\% & 0.243430 \\  \hline
          \multicolumn{3}{ |c| }{passive deep learning (PDL) benchmark} & 100\% & 0\% & 50\% & 0.712441 & 80\% & 0\% & 27\% & 0223089 \\  \hline
          \multicolumn{3}{ |c| }{random forest (RF) baseline} & 0\% & 0\% & 0\% & 1.411788 & 0\% & 0\% & 0\% & 0.304446 \\  \hline
          \multicolumn{1}{ |c }{Spatio-temporal}& $\mathbf{x}_{st}$ & rnd $d_{c, st}$ & 100\% & 100\% & 77\% & 0.438415 & 54\% & 100\% & 71\% & 0.511841 \\ 
          \multicolumn{1}{ |c }{ }& & max $d_{c, st}$ & 100\% & 100\% & 74\% & 0.500414 & 18\% & 100\% & 60\% & 0.703498 \\ 
          \multicolumn{1}{ |c }{ }& & min $d_{c, st}$ & 100\% & 100\% & 78\% & 0.428802 & 20\% & 100\% & 63\% & 0.648593 \\ 
          \multicolumn{1}{ |c }{ }& & avg $d_{c, st}$ & 100\% & 100\% & 76\% & 0.472252 & 18\% & 100\% & 57\% & 0.752075 \\ 
          \multicolumn{1}{ |c }{ }& $\mathbf{x}_{t,s}$ & rnd $d_{c, (t,s)}$ & 100\% & 100\% & 87\% & 0.259112 & 75\% & 100\% & 70\% & 0.524477 \\ 
          \multicolumn{1}{ |c }{ }& & max $d_{c, (t,s)}$ & 100\% & 100\% & 89\% & 0.218669 & 58\% & 100\% & 56\% & 0.771426 \\ 
          \multicolumn{1}{ |c }{ }& & min $d_{c, (t,s)}$ & 100\% & 100\% & 90\% & 0.197869 & 72\% & 100\% & 76\% & 0.417253 \\ 
          \multicolumn{1}{ |c }{ }& & avg $d_{c, (t,s)}$ & 100\% & 100\% & 86\% & 0.273082 & 66\% & 100\% & 69\% & 0.537644 \\ 
          \multicolumn{1}{ |c }{ }& $\mathbf{\hat{y}}_{t,s}$ & rnd $d_{c, (t,s)}$ & 100\% & 100\% & 96\% & 0.069790 & 47\% & 100\% & 88\% & 0.211104 \\ 
          \multicolumn{1}{ |c }{ }& & max $d_{c, (t,s)}$ & 100\% & 100\% & 97\% & 0.067190 & 40\% & 97\% & 85\% & 0.266376 \\ 
          \multicolumn{1}{ |c }{ }& & min $d_{c, (t,s)}$ & 100\% & 100\% & 96\% & 0.072054 & 48\% & 100\% & 87\% & 0.232354 \\ 
          \multicolumn{1}{ |c }{ }& & avg $d_{c, (t,s)}$ & 100\% & 100\% & 96\% & 0.068377 & 44\% & 100\% & 86\% & 0.243160 \\ 
          \multicolumn{1}{ |c }{ }& $\mathbf{y}_{t,s}$ & rnd $d_{c, (t,s)}$ & 100\% & 100\% & 97\% & 0.062084 & 39\% & 100\% & 72\% & 0.492224 \\ 
          \multicolumn{1}{ |c }{ }& & max $d_{c, (t,s)}$ & 100\% & 95\% & 95\% & 0.103196 & 22\% & 66\% & 47\% & 0.926593 \\ 
          \multicolumn{1}{ |c }{ }& & min $d_{c, (t,s)}$ & 100\% & 100\% & 96\% & 0.082361 & 25\% & 86\% & 68\% & 0.563649 \\ 
          \multicolumn{1}{ |c }{ }& & avg $d_{c, (t,s)}$ & 100\% & 98\% & 96\% & 0.071590 & 24\% & 77\% & 57\% & 0.760949 \\   \hline
          \multicolumn{3}{ |c| }{passive deep learning (PDL) benchmark} & 100\% & 100\% & 53\% & 0.912378 & 80\% & 100\% & 45\% & 0.963880 \\  \hline
          \multicolumn{3}{ |c| }{random forest (RF) baseline} & 0\% & 0\% & 0\% & 1.936088 & 0\% & 0\% & 0\% & 1.757695 \\ 
         \hline
        \end{tabular}
    }
    \end{center}
\end{table}


\begin{thebibliography}{1}
\expandafter\ifx\csname url\endcsname\relax
  \def\url#1{\texttt{#1}}\fi
\expandafter\ifx\csname urlprefix\endcsname\relax\def\urlprefix{URL }\fi
\providecommand{\bibinfo}[2]{#2}
\providecommand{\eprint}[2][]{\url{#2}}

  
  
  
\bibitem{Patt2015}
\bibinfo{author}{Patt, A.} 
\newblock \bibinfo{title}{{Transforming Energy - Solving Climate Change with Technology Policy}}.
\newblock \emph{\bibinfo{journal}{Cambridge Press}}
  (\bibinfo{year}{2015}).
  
\bibitem{IPCC2018}
\bibinfo{author}{Rogelj, J.} \& et al.
\newblock \bibinfo{title}{{Mitigation Pathways Compatible with 1.5°C in the Context of Sustainable Development}}.
\newblock \emph{\bibinfo{journal}{Global Warming of 1.5°C. An IPCC Special Report on the impacts of global warming of 1.5°C above pre-industrial levels and related global greenhouse gas emission pathways, in the context of strengthening the global response to the threat of climate change, sustainable development, and efforts to eradicate poverty}}
  (\bibinfo{year}{2018}).
  
\bibitem{IPCC2021}
\bibinfo{author}{Masson-Delmotte, V.} \& et al.
\newblock \bibinfo{title}{{Climate Change 2021: The Physical Science Basis. Contribution of Working Group I to the Sixth Assessment Report of the Intergovernmental Panel on Climate Change}}.
\newblock \emph{\bibinfo{journal}{IPCC}}
  (\bibinfo{year}{2021}).

\bibitem{Hahn2009}
\bibinfo{author}{Hahn, H.}, \bibinfo{author}{Meyer-Nieberg, S.} \& \bibinfo{author}{Pickl, S.}
\newblock \bibinfo{title}{{Electric load forecasting methods: Tools for decision making}}.
\newblock \emph{\bibinfo{journal}{European Journal of Operational Research}}
  \textbf{\bibinfo{volume}{199}}, \bibinfo{pages}{}
  \bibinfo{URL}{https://doi.org/10.1016/j.ejor.2009.01.062}
  (\bibinfo{year}{2009}).
  
\bibitem{Soliman2010}
\bibinfo{author}{Soliman, A.-h. S.} \& \bibinfo{author}{Al-Kandari, A. M.}
\newblock \bibinfo{title}{{Electric Load Forecasting}}.
\newblock \emph{\bibinfo{journal}{Butterworth-Heinemann}}
  \textbf{\bibinfo{volume}{}}, \bibinfo{pages}{}
  \bibinfo{URL}{https://doi.org/10.1016/C2009-0-60996-X}
  (\bibinfo{year}{2010}).
  
\bibitem{Alfares2002}
\bibinfo{author}{Alfares, H. K.} \& \bibinfo{author}{Nazeeruddin, M.}
\newblock \bibinfo{title}{{Electric load forecasting: literature survey and classification of methods}}.
\newblock \emph{\bibinfo{journal}{International Journal of Systems Science}}
  \textbf{\bibinfo{volume}{33}}, \bibinfo{pages}{23-24}
  \bibinfo{URL}{10.1080/00207720110067421}
  (\bibinfo{year}{2002}).
  
\bibitem{Nti2020}
\bibinfo{author}{Kofi Nti, I.}, \bibinfo{author}{Teimeh, M.},
\bibinfo{author}{Nyarko-Boateng, O.} \& \bibinfo{author}{Adekoya, A. F.}
\newblock \bibinfo{title}{{Electricity load forecasting: a systematic review}}.
\newblock \emph{\bibinfo{journal}{Journal of Electrical Systems and Information Technology}}
  \textbf{\bibinfo{volume}{7}}, \bibinfo{pages}{}
  (\bibinfo{year}{2020}).

\bibitem{Shi2017}
\bibinfo{author}{Shi, J.}, \bibinfo{author}{Liu, Y.} \& \bibinfo{author}{Yu, N.}
\newblock \bibinfo{title}{{Spatio-temporal modeling of electric loads}}.
\newblock \emph{\bibinfo{journal}{IEEE}}
  \bibinfo{URL}{10.1109/NAPS.2017.8107311}
  (\bibinfo{year}{2017}).

\bibitem{Tascikaraoglu2018}
\bibinfo{author}{Tascikaraoglu, A.}
\newblock \bibinfo{title}{{Evaluation of spatio-temporal forecasting methods in various smart city applications}}.
\newblock \emph{\bibinfo{journal}{Renewable and Sustainable Energy Reviews}}
  \bibinfo{URL}{https://doi.org/10.1016/j.rser.2017.09.078}
  \textbf{\bibinfo{volume}{82}}, \bibinfo{pages}{424-435}
  (\bibinfo{year}{2018}).
  
\bibitem{Severiano2021}
\bibinfo{author}{Severiano, C. A.}, \bibinfo{author}{Cândido de Lima eSilva, P.},
\bibinfo{author}{Cohen, M. W.} \& \bibinfo{author}{Gadelha Guimarãesae, F.}
\newblock \bibinfo{title}{{Evolving fuzzy time series for spatio-temporal forecasting in renewable energy systems}}.
\newblock \emph{\bibinfo{journal}{Renewable Energy}}
  \bibinfo{URL}{https://doi.org/10.1016/j.renene.2021.02.117}
  \textbf{\bibinfo{volume}{171}}, \bibinfo{pages}{764-783}
  (\bibinfo{year}{2021}).

\bibitem{Willis2002}
\bibinfo{author}{Willis, H. L.}
\newblock \bibinfo{title}{{Spatial Electric Load Forecasting}}.
  (\bibinfo{year}{2002}).
  
\bibitem{Rolf2021}
\bibinfo{author}{Rolf, E.} \& et al.
\newblock \bibinfo{title}{{A generalizable and accessible approach to machine learning with global satellite imagery}}.
\newblock \emph{\bibinfo{journal}{Nature Communications}}
  \textbf{\bibinfo{volume}{12}}, \bibinfo{pages}{}
  (\bibinfo{year}{2021}).
  
\bibitem{Burke2021}
\bibinfo{author}{Burke, M.}, \bibinfo{author}{Driscoll, A.} 
\bibinfo{author}{Lobell, D. B.} \& \bibinfo{author}{Ermon, S.} 
\newblock \bibinfo{title}{{Using satellite imagery to understand and promote
sustainable development}}.
\newblock \emph{\bibinfo{journal}{Science}}
  \bibinfo{URL}{10.1126/science.abe8628}
  \textbf{\bibinfo{volume}{371}}, \bibinfo{pages}{}
  (\bibinfo{year}{2021}).   
  
\bibitem{Melo2014}
\bibinfo{author}{Melo, J. D.} \& \bibinfo{author}{Carreno, E. M.}
\newblock \bibinfo{title}{{Data Issues in Spatial Electric Load Forecasting}}.
\newblock \emph{\bibinfo{journal}{IEEE}}
  (\bibinfo{year}{2014}). 

\bibitem{Milam2014}
\bibinfo{author}{Milam, M.} \& \bibinfo{author}{Venayagamoorthy, G. K.}
\newblock \bibinfo{title}{{Smart meter deployment: US initiatives}}.
\newblock \emph{\bibinfo{journal}{IEEE}}
  \bibinfo{URL}{10.1109/ISGT.2014.6816507}
  (\bibinfo{year}{2014}). 

\bibitem{Sovacool2021}
\bibinfo{author}{Sovacool, B. K.}, \bibinfo{author}{Hook, A.},
\bibinfo{author}{Sareen, S.} \& \bibinfo{author}{Geels, F. W.}
\newblock \bibinfo{title}{{Global sustainability, innovation and governance dynamics of national smart electricity meter transitions}}.
\newblock \emph{\bibinfo{journal}{Global Environmental Change}}
  \textbf{\bibinfo{volume}{68}}, \bibinfo{pages}{}
  \bibinfo{URL}{10.1016/j.gloenvcha.2021.102272}
  (\bibinfo{year}{2021}).

\bibitem{Kezunovic2013}
\bibinfo{author}{Kezunovic, M.}, \bibinfo{author}{Xie, L.} \& 
\bibinfo{author}{Grijalva, S.}
\newblock \bibinfo{title}{{The Role of Big Data in Improving Power System Operation and Protection}}.
\newblock \emph{\bibinfo{journal}{IEEE}}
  \bibinfo{URL}{10.1109/IREP.2013.6629368}
  (\bibinfo{year}{2013}).

\bibitem{Yu2015}
\bibinfo{author}{Yu, N.} \& et al.
\newblock \bibinfo{title}{{The Role of Big Data in Improving Power System Operation and Protection}}.
\newblock \emph{\bibinfo{journal}{IEEE}}
  \bibinfo{URL}{10.1109/ISGT.2015.7131868}
  (\bibinfo{year}{2015}).
  
\bibitem{Stein2020}
\bibinfo{author}{Stein, A. L} 
\newblock \bibinfo{title}{{Artificial Intelligence and Climate Change}}.
\newblock \emph{\bibinfo{journal}{Yale Journal on Regulation}}
  \textbf{\bibinfo{volume}{37}}, \bibinfo{pages}{890-934}
  (\bibinfo{year}{2020}).

\bibitem{Rolnick2019}
\bibinfo{author}{Rolnick, D.} \& et al.
\newblock \bibinfo{title}{{Tackling Climate Change with Machine Learning}}.
\newblock \bibinfo{URL}{Preprint at https://arxiv.org/abs/1906.05433}
  (\bibinfo{year}{2019}).


\bibitem{Settles2010}
\bibinfo{author}{Settles, B.}
\newblock \bibinfo{title}{{Active Learning Literature Survey}}.
\newblock \emph{\bibinfo{journal}{Computer Sciences Technical Report}}
  \textbf{\bibinfo{volume}{1648}}, \bibinfo{pages}{}
  \bibinfo{URL}{http://digital.library.wisc.edu/1793/60660}
  (\bibinfo{year}{2010}).  

\bibitem{Dasgupta2011}
\bibinfo{author}{Dasgupta, S.}
\newblock \bibinfo{title}{{Two faces of active learning}}.
\newblock \emph{\bibinfo{journal}{Theoretical Computer Science}}
  \textbf{\bibinfo{volume}{412}}, \bibinfo{pages}{1767–1781}
  \bibinfo{URL}{https://doi.org/10.1016/j.tcs.2010.12.054}
  (\bibinfo{year}{2011}).
  
\bibitem{Kumar2020}
\bibinfo{author}{Kumar, P.} \& \bibinfo{author}{Gupta, A.}
\newblock \bibinfo{title}{{Active Learning Query Strategies for Classification, Regression and Clustering: A Survey}}.
\newblock \emph{\bibinfo{journal}{Journal of Computer Science and Technology}}
  \textbf{\bibinfo{volume}{35}}, \bibinfo{pages}{913-945}
  \bibinfo{URL}{https://doi.org/10.1007/s11390-020-9487-4}
  (\bibinfo{year}{2020}).

\bibitem{Kuo2015}
\bibinfo{author}{Kuo, P.}, \bibinfo{author}{Liang, D.},
\bibinfo{author}{Gao, L.} \& \bibinfo{author}{Lou, J.}
\newblock \bibinfo{title}{{Probabilistic electricity price forecasting with variational heteroscedastic Gaussian process and active learning}}.
\newblock \emph{\bibinfo{journal}{Energy Conversion and Management}}
  \textbf{\bibinfo{volume}{89}}, \bibinfo{pages}{298-308}
  \bibinfo{URL}{https://doi.org/10.1016/j.enconman.2014.10.003}
  (\bibinfo{year}{2015}).

\bibitem{Wang2019}
\bibinfo{author}{Wang, Z.}, \bibinfo{author}{Zhao, B.},
\bibinfo{author}{Guo, H.}, \bibinfo{author}{Tang, L.} \&
\bibinfo{author}{Peng, Y.}
\newblock \bibinfo{title}{{Deep Ensemble Learning Model for Short-Term Load Forecasting within Active Learning Framework }}.
\newblock \emph{\bibinfo{journal}{Energies}}
  \textbf{\bibinfo{volume}{12}}, \bibinfo{pages}{}
  \bibinfo{URL}{https://doi.org/10.3390/en12203809}
  (\bibinfo{year}{2019}).

\bibitem{Zhang2021}
\bibinfo{author}{Zhang, L.} \& \bibinfo{author}{Wen, J.}
\newblock \bibinfo{title}{{Active learning strategy for high fidelity short-term data-driven building energy forecasting}}.
\newblock \emph{\bibinfo{journal}{Energy and Buildings}}
  \textbf{\bibinfo{volume}{244}}, \bibinfo{pages}{}
  \bibinfo{URL}{https://doi.org/10.1016/j.enbuild.2021.111026}
  (\bibinfo{year}{2021}).


  
\bibitem{Siddiqui2020}
\bibinfo{author}{Siddiqui, Y.}, \bibinfo{author}{Valentin, J.} \& 
\bibinfo{author}{Nießner, M.}
\newblock \bibinfo{title}{{ViewAL: Active Learning with Viewpoint Entropy for Semantic Segmentation}}.
\newblock \emph{\bibinfo{journal}{CVPR}}
  (\bibinfo{year}{2020}).
  
\bibitem{Chen2020}
\bibinfo{author}{Chen, T.}, \& et al.
\newblock \bibinfo{title}{{Big Self-Supervised Models are Strong Semi-Supervised Learners}}.
\newblock \emph{\bibinfo{journal}{NeurIPS}}
  \textbf{\bibinfo{volume}{}}, \bibinfo{pages}{}
  (\bibinfo{year}{2020}).
  
\bibitem{Yue2021}
\bibinfo{author}{Yue, X.}, \& et al.
\newblock \bibinfo{title}{{Prototypical Cross-domain Self-supervised Learning for Few-shot Unsupervised Domain Adaptation}}.
\newblock \emph{\bibinfo{journal}{CVPR}}
  \textbf{\bibinfo{volume}{}}, \bibinfo{pages}{}
  (\bibinfo{year}{2021}).

\bibitem{Yang2020}
\bibinfo{author}{Yang, W.}, \& et al.
\newblock \bibinfo{title}{{Class Distribution Alignment for Adversarial Domain Adaptation}}.
\newblock \emph{\bibinfo{journal}{CVPR}}
  \textbf{\bibinfo{volume}{}}, \bibinfo{pages}{}
  (\bibinfo{year}{2020}).

\bibitem{Saito2020}
\bibinfo{author}{Saito, K.}, \bibinfo{author}{Saenko, K.} \&
\bibinfo{author}{Liu, M.-Y.}
\newblock \bibinfo{title}{{COCO-FUNIT: Few-Shot Unsupervised Image Translation with a Content Conditioned Style Encoder}}.
\newblock \emph{\bibinfo{journal}{CVPR}}
  \textbf{\bibinfo{volume}{}}, \bibinfo{pages}{}
  (\bibinfo{year}{2020}).

\bibitem{Jain2016}
\bibinfo{author}{Jain, A.}, \bibinfo{author}{Zamir, A. R.}, \bibinfo{author}{Savarese, S.} \& \bibinfo{author}{Saxena, A.}
\newblock \bibinfo{title}{{Structural-RNN: Deep Learning on Spatio-Temporal Graphs}}.
\newblock \emph{\bibinfo{journal}{CVPR}}
  \textbf{\bibinfo{volume}{}}, \bibinfo{pages}{}
  (\bibinfo{year}{2016}).



\bibitem{Mikolov2013}
\bibinfo{author}{Mikolov, T.}, \bibinfo{author}{Chen, K.}, \bibinfo{author}{Corrado, G.} \& \bibinfo{author}{Dean, J.}
\newblock \bibinfo{title}{{Efficient Estimation of Word Representations in Vector Space}}.
\newblock \bibinfo{URL}{Preprint at https://arxiv.org/abs/1301.3781}
  (\bibinfo{year}{2013}).
  
\bibitem{Frome2013}
\bibinfo{author}{Frome, A.} \& \bibinfo{author}{et al.}
\newblock \bibinfo{title}{{DeViSE: A Deep Visual-Semantic Embedding Model}}.
\newblock \emph{\bibinfo{journal}{NIPS}}
  \textbf{\bibinfo{volume}{}}, \bibinfo{pages}{2121–2129}
  (\bibinfo{year}{2013}).
  
\bibitem{Pennington2014}
\bibinfo{author}{Pennington, J.}, \bibinfo{author}{Socher, R.} \& \bibinfo{author}{Manning, C. D.}
\newblock \bibinfo{title}{{GloVe: Global Vectors for Word Representation}}.
\newblock \emph{\bibinfo{journal}{EMNLP}}
\newblock \bibinfo{URL}{https://doi.org/10.3115/v1/D14-1162}
  (\bibinfo{year}{2014}).
  
\bibitem{Perozzi2014}
\bibinfo{author}{Perozzi, B.}, \bibinfo{author}{Al-Rfou, R.} \& \bibinfo{author}{Skiena, S.}
\newblock \bibinfo{title}{{DeepWalk: Online Learning of Social Represenations}}.
  \bibinfo{URL}{Preprint at https://arxiv.org/abs/1403.6652}
  \bibinfo{URL}{https://doi.org/10.1145/2623330.2623732}
  (\bibinfo{year}{2014}).
  
\bibitem{Devlin2019}
\bibinfo{author}{Devlin, J.}, \bibinfo{author}{Chang, M.-W.},
\bibinfo{author}{Lee, K.} \& \bibinfo{author}{Toutanova, K.}
\newblock \bibinfo{title}{{BERT: Pre-training of Deep Bidirectional Transformers for Language Understanding}}.
\newblock \bibinfo{URL}{Preprint at https://arxiv.org/abs/1810.04805}
  (\bibinfo{year}{2019}).

\bibitem{GeoVITE2020}
\bibinfo{author}{Federal Office of Topography}
\newblock \bibinfo{title}{{GeoVITE - User-friendly Geodata Service}}.
  \bibinfo{URL}{https://geovite.ethz.ch/},
  (\bibinfo{year}{2020}).

\bibitem{Pfenninger2016}
\bibinfo{author}{Pfenninger, S.} \& \bibinfo{author}{Staffel, I.}
\newblock \bibinfo{title}{{Long-term patterns of European PV output using 30 years of validated hourly reanalysis and satellite data}}.
\newblock \emph{\bibinfo{journal}{Energy}}
  \textbf{\bibinfo{volume}{114}}, \bibinfo{pages}{1251-1265}
  (\bibinfo{year}{2016}).
  
\bibitem{Staffel2016}
\bibinfo{author}{Staffel, I.} \& \bibinfo{author}{Pfenninger, S.}
\newblock \bibinfo{title}{{Using Bias-Corected Reanalysis to Simulate Current and Future Wind Power Output}}.
\newblock \emph{\bibinfo{journal}{Energy}}
  \textbf{\bibinfo{volume}{114}}, \bibinfo{pages}{1224-1239}
  (\bibinfo{year}{2016}).

\bibitem{Ash2020}
\bibinfo{author}{Ash, J. T.}, \bibinfo{author}{Zhang, C.},
\bibinfo{author}{Krishnamurthy, A.}, \bibinfo{author}{Langford, J.} \& 
\bibinfo{author}{Agarwal, A.}
\newblock \bibinfo{title}{{Deep Batch Active Learning by Diverse, Uncertain Gradient Lower Bounds}}.
\newblock \emph{\bibinfo{journal}{ICLR}}
  (\bibinfo{year}{2020}).

\bibitem{RenewablesNinja2020}
\bibinfo{author}{Pfenninger, S.} \& \bibinfo{author}{Staffel, I.}
  \bibinfo{URL}{https://www.renewables.ninja},
  (\bibinfo{year}{2020}).

\bibitem{Aryandoust2021}
\bibinfo{author}{Aryandoust, A.}
\newblock \bibinfo{title}{{Enhanced spatio-temporal electric load forecasts using less data with active deep learning}}.
  \bibinfo{URL}{https://github.com/ArsamAryandoust/DataSelectionMaps},
  (\bibinfo{year}{2021}).

\bibitem{Aryandoust2021_2}
\bibinfo{author}{Aryandoust, A.}
\newblock \bibinfo{title}{{Enhanced spatio-temporal electric load forecasts using less data with active deep learning}}.
  \bibinfo{URL}{https://pypi.org/project/altility},
  (\bibinfo{year}{2021}).

\bibitem{Aryandoust2021_3}
\bibinfo{author}{Aryandoust, A.}
\newblock \bibinfo{title}{{Enhanced spatio-temporal electric load forecasts using less data with active deep learning}}.
  \bibinfo{URL}{https://hub.docker.com/r/aryandoustarsam/altility},
  (\bibinfo{year}{2021}).



\bibitem{LeCun2015}
\bibinfo{author}{LeCun, Y.}, \bibinfo{author}{Bengio, Y.} \& 
\bibinfo{author}{Hinton, G.}
\newblock \bibinfo{title}{{Deep learning}}.
\newblock \emph{\bibinfo{journal}{Nature}}
  \textbf{\bibinfo{volume}{521}}, \bibinfo{pages}{436–444}
  (\bibinfo{year}{2015}).
  
\bibitem{Schrittwieser2020}
\bibinfo{author}{Schrittwieser, J.} \& et al.
\newblock \bibinfo{title}{{Mastering Atari, Go, chess and shogi by planning with a learned model}}.
\newblock \emph{\bibinfo{journal}{Nature}}
  \textbf{\bibinfo{volume}{588}}, \bibinfo{pages}{604-609}
  (\bibinfo{year}{2020}).
  
\bibitem{Hejazi2018}
\bibinfo{author}{Akhavan-Hejazi, H.} \& \bibinfo{author}{Mohsenian-Rad, H.}
\newblock \bibinfo{title}{{Power systems big data analytics: An assessment of paradigm shift barriers and prospects}}.
\newblock \emph{\bibinfo{journal}{Energy Reports}}
  \textbf{\bibinfo{volume}{4}}, \bibinfo{pages}{91-100}
  \bibinfo{URL}{https://doi.org/10.1016/j.egyr.2017.11.002}
  (\bibinfo{year}{2018}).
  
\bibitem{Shahra2020}
\bibinfo{author}{Shahra, E. Q.} \& \bibinfo{author}{Wu, W.}
\newblock \bibinfo{title}{{Water contaminants detection using sensor placement approach in smart water networks}}.
\newblock \emph{\bibinfo{journal}{Journal of Ambient Intelligence and Humanized Computing }}
  \textbf{\bibinfo{volume}{}}, \bibinfo{pages}{}
  (\bibinfo{year}{2020}).  

\bibitem{Hong2015}
\bibinfo{author}{Hong, T.}, \bibinfo{author}{Wang, P.} \& 
\bibinfo{author}{White, L.}
\newblock \bibinfo{title}{{Weather station selection for electric load forecasting}}.
\newblock \emph{\bibinfo{journal}{International Journal of Forecasting}}
  \textbf{\bibinfo{volume}{31}}, \bibinfo{pages}{286-295}
  \bibinfo{URL}{https://doi.org/10.1016/j.ijforecast.2014.07.001}
  (\bibinfo{year}{2015}).

\bibitem{Lewis1994_2}
\bibinfo{author}{Lewis, D. D.} \&  \bibinfo{author}{Catlett, J.}
\newblock \bibinfo{title}{{Heterogeneous Uncertainty Sampling for Supervised Learning}}.
\newblock \emph{\bibinfo{journal}{Machine Learning Proceedings}}
  \textbf{\bibinfo{volume}{11}}, \bibinfo{pages}{148-156}
  \bibinfo{URL}{https://doi.org/10.1016/B978-1-55860-335-6.50026-X}
  (\bibinfo{year}{1994}).
  
\bibitem{Lindenbaum2004}
\bibinfo{author}{Lindenbaum, M.}, \bibinfo{author}{Markovitch, S.}
\& \bibinfo{author}{Rusakov, D.}
\newblock \bibinfo{title}{{Selective sampling for nearest neighbor classifiers}}.
\newblock \emph{\bibinfo{journal}{Machine Learning}}
  \textbf{\bibinfo{volume}{54}}, \bibinfo{pages}{125-152}
  (\bibinfo{year}{2004}).

\bibitem{Culotta2005}
\bibinfo{author}{Culotta, A.} \& \bibinfo{author}{McCallum, A.}
\newblock \bibinfo{title}{{Reducing labeling effort for structured prediction tasks}}.
\newblock \emph{\bibinfo{journal}{AAAI}}
  \textbf{\bibinfo{volume}{2}}, \bibinfo{pages}{746-751}
  (\bibinfo{year}{2005}).
  
\bibitem{Koerner2006}
\bibinfo{author}{Körner, C.} \& \bibinfo{author}{Wrobel, S.}
\newblock \bibinfo{title}{{Multi-class ensemble-based active learning}}.
\newblock \emph{\bibinfo{journal}{ECML}}
  \textbf{\bibinfo{volume}{}}, \bibinfo{pages}{687-694}
  (\bibinfo{year}{2006}).
  
\bibitem{Schein2007}
\bibinfo{author}{Schein, A. I.} \& \bibinfo{author}{Ungar, L. H.}
\newblock \bibinfo{title}{{Active learning for logistic regression: An evaluation}}.
\newblock \emph{\bibinfo{journal}{Machine Learning}}
  \textbf{\bibinfo{volume}{68 (3)}}, \bibinfo{pages}{235-265}
  \bibinfo{URL}{https://doi.org/10.1007/s10994-007-5019-5}
  (\bibinfo{year}{2007}).

\bibitem{Dagan1995}
\bibinfo{author}{Dagan, I.} \& \bibinfo{author}{Engelson, S.}
\newblock \bibinfo{title}{{Committee-based sampling for training probabilistic classifiers}}.
\newblock \emph{\bibinfo{journal}{ICML}}
  \textbf{\bibinfo{volume}{}}, \bibinfo{pages}{150-157}
  \bibinfo{URL}{https://doi.org/10.1016/B978-1-55860-377-6.50027-X}
  (\bibinfo{year}{1995}).
  
\bibitem{McCallum1998}
\bibinfo{author}{McCallum, A.} \& \bibinfo{author}{Nigam, K.}
\newblock \bibinfo{title}{{Employing EM in pool-based active learning for text classification}}.
\newblock \emph{\bibinfo{journal}{ICML}}
  \textbf{\bibinfo{volume}{}}, \bibinfo{pages}{359-367}
  (\bibinfo{year}{1998}).  

\bibitem{Muslea2000}
\bibinfo{author}{Muslea, I.}, \bibinfo{author}{Minton, S.} \& 
\bibinfo{author}{Knoblock, C. A.}
\newblock \bibinfo{title}{{Selective sampling with redundant views}}.
\newblock \emph{\bibinfo{journal}{AAAI}}
  \textbf{\bibinfo{volume}{}}, \bibinfo{pages}{621-626}
  (\bibinfo{year}{2000}).  
  
\bibitem{Melville2004}
\bibinfo{author}{Melville, P.} \& \bibinfo{author}{Mooney, R. J.}
\newblock \bibinfo{title}{{Diverse ensembles for active learning}}.
\newblock \emph{\bibinfo{journal}{ICML}}
  \textbf{\bibinfo{volume}{}}, \bibinfo{pages}{584-591}
  \bibinfo{URL}{https://doi.org/10.1145/1015330.1015385}
  (\bibinfo{year}{2004}).
  
\bibitem{Hanneke2014}
\bibinfo{author}{Hanneke, S.}
\newblock \bibinfo{title}{{Theory of Disagreement-Based Active Learning}}.
\newblock \emph{\bibinfo{journal}{Foundations and Trends in Machine Learning}}
  \textbf{\bibinfo{volume}{7}}, \bibinfo{pages}{31–309}
  \bibinfo{URL}{https://doi.org/10.1561/2200000037}
  (\bibinfo{year}{2014}). 

\bibitem{MacKay1992}
\bibinfo{author}{MacKay, D. J. C.}
\newblock \bibinfo{title}{{Information-based objective functions for active data selection}}.
\newblock \emph{\bibinfo{journal}{Neural Computation}}
  \textbf{\bibinfo{volume}{4}}, \bibinfo{pages}{590-604}
  \bibinfo{URL}{http://dx.doi.org/10.1162/neco.1992.4.4.590}
  (\bibinfo{year}{1992}). 
  
\bibitem{Cohn1996}
\bibinfo{author}{Cohn, D. A.}, \bibinfo{author}{Ghahramani, Z.} \& 
\bibinfo{author}{Jordan, M. I.}
\newblock \bibinfo{title}{{Active learning with statistical models}}.
\newblock \emph{\bibinfo{journal}{Journal of Artificial Intelligence Research}}
  \textbf{\bibinfo{volume}{}}, \bibinfo{pages}{129-145}
  (\bibinfo{year}{1996}). 

\bibitem{Zhang2000}
\bibinfo{author}{Zhang, T.} \& \bibinfo{author}{Oles, F. J.}
\newblock \bibinfo{title}{{A probability analysis on the value of unlabeled data for classification problems}}.
\newblock \emph{\bibinfo{journal}{ICML}}
  \textbf{\bibinfo{volume}{}}, \bibinfo{pages}{1191-1198}
  (\bibinfo{year}{2000}).
  
\bibitem{Hoi2006}
\bibinfo{author}{Hoi, S. C. H.}, \bibinfo{author}{Jin, R.} \& 
\bibinfo{author}{Lyu, M. R.}
\newblock \bibinfo{title}{{Large-scale text categorization by batch mode active learning}}.
\newblock \emph{\bibinfo{journal}{WWW}}
  \textbf{\bibinfo{volume}{}}, \bibinfo{pages}{633-642}
  \bibinfo{URL}{https://doi.org/10.1145/1135777.1135870}
  (\bibinfo{year}{2006}).

\bibitem{Roy2001}
\bibinfo{author}{Roy, N.} \& \bibinfo{author}{McCallum, A.}
\newblock \bibinfo{title}{{Toward optimal active learning through sampling estimation of error reduction}}.
\newblock \emph{\bibinfo{journal}{ICML}}
  \textbf{\bibinfo{volume}{}}, \bibinfo{pages}{441-448}
  (\bibinfo{year}{2001}). 

\bibitem{Zhu2003}
\bibinfo{author}{Zhu, X.}, \& \bibinfo{author}{Lafferty, J.}
\bibinfo{author}{Ghahramani, Z.}
\newblock \bibinfo{title}{{Combining Active Learning and Semi-Supervised Learning Using Gaussian Fields and Harmonic Functions}}.
\newblock \emph{\bibinfo{journal}{Proceedings of the ICML Workshop on Continuum from Labeled to Unlabeled Data}}
  \textbf{\bibinfo{volume}{}}, \bibinfo{pages}{58-65}
  (\bibinfo{year}{2003}). 

\bibitem{Guo2007}
\bibinfo{author}{Guo, Y.} \& \bibinfo{author}{Greiner, R.}
\newblock \bibinfo{title}{{Optimistic active learning using mutual information}}.
\newblock \emph{\bibinfo{journal}{IJCAI}}
  \textbf{\bibinfo{volume}{}}, \bibinfo{pages}{823-829}
  (\bibinfo{year}{2007}).

\bibitem{Moskovitch2007}
\bibinfo{author}{Moskovitch, R.} \& \bibinfo{author}{et al.}
\newblock \bibinfo{title}{{Improving the detection of unknown computer worms activity using active learning}}.
\newblock \emph{\bibinfo{journal}{KI}}
  (\bibinfo{year}{2007}).

\bibitem{Settles2007}
\bibinfo{author}{Settles, B.}, \bibinfo{author}{Craven, M.} \& 
\bibinfo{author}{Ray, S.}
\newblock \bibinfo{title}{{Multiple-instance active learning}}.
\newblock \emph{\bibinfo{journal}{NIPS}}
  \textbf{\bibinfo{volume}{}}, \bibinfo{pages}{1289–1296}
  (\bibinfo{year}{2007}).
  
\bibitem{Settles2008}
\bibinfo{author}{Settles, B.} \& \bibinfo{author}{Craven, M.}
\newblock \bibinfo{title}{{An analysis of active learning strategies for sequence labeling tasks}}.
\newblock \emph{\bibinfo{journal}{EMNLP}}
  \textbf{\bibinfo{volume}{}}, \bibinfo{pages}{1070–1079}
  \bibinfo{URL}{https://doi.org/10.3115/1613715.1613855}
  (\bibinfo{year}{2008}).  

\bibitem{Nguyen2004}
\bibinfo{author}{Nguyen, H. T.} \& \bibinfo{author}{Smeulders, A.}
\newblock \bibinfo{title}{{Active learning using pre-clustering}}.
\newblock \emph{\bibinfo{journal}{ICML}}
  \bibinfo{URL}{https://doi.org/10.1145/1015330.1015349}
  (\bibinfo{year}{2004}).

\bibitem{Xu2007}
\bibinfo{author}{Xu, Z.}, \bibinfo{author}{Akella, R.} \& 
\bibinfo{author}{Zhang, Y.}
\newblock \bibinfo{title}{{Incorporating diversity and density in active learning for relevance feedback}}.
\newblock \emph{\bibinfo{journal}{ECIR}}
  \textbf{\bibinfo{volume}{}}, \bibinfo{pages}{246–257}
  (\bibinfo{year}{2007}).
  
\bibitem{Dasgupta2008}
\bibinfo{author}{Dasgupta, S.} \& \bibinfo{author}{Hsu, D. J.}
\newblock \bibinfo{title}{{Hierarchical sampling for active learning}}.
\newblock \emph{\bibinfo{journal}{ICML}}
  \bibinfo{URL}{https://doi.org/10.1145/1390156.1390183}
  (\bibinfo{year}{2008}).

\bibitem{Gonzalez2016}
\bibinfo{author}{Gonzalez, J. A.}, \bibinfo{author}{Rodriguez-Cortes, F. J.},
\bibinfo{author}{Cronie, O.} \& \bibinfo{author}{Mateu, J.}
\newblock \bibinfo{title}{{Spatio-temporal point process statistics: A review}}.
\newblock \emph{\bibinfo{journal}{Spatial Statistics}}
  \textbf{\bibinfo{volume}{18 (B)}}, \bibinfo{pages}{505-544}
  (\bibinfo{year}{2016}).


\end{thebibliography}
\end{document}